\documentclass[journal=jacsat,manuscript=article]{achemso}

\usepackage{chemformula} % Formula subscripts using \ch{}
\usepackage[T1]{fontenc} % Use modern font encodings
\usepackage{multirow}
\usepackage{makecell}
\usepackage{graphicx}
\usepackage{adjustbox}
\usepackage{booktabs}
\usepackage{rotating}
\usepackage{seqsplit} 
\usepackage[numbers]{natbib}
\usepackage{hyperref}
\usepackage{longtable}
\usepackage{seqsplit}
\usepackage{fancyhdr}
\mciteErrorOnUnknownfalse

\newcommand{\beginsupplement}{
	\setcounter{table}{0}
	\renewcommand{\thetable}{S\arabic{table}}
	\renewcommand{\theHtable}{\thetable}
	\setcounter{figure}{0}
	\renewcommand{\thefigure}{S\arabic{figure}}
	\renewcommand{\theHfigure}{\thefigure}
	\setcounter{page}{1}
	\pagestyle{fancy}
	\fancyhf{}
	\cfoot{S\thepage}
	\renewcommand{\headrulewidth}{0pt}
}

\author{Yifan Deng}
\affiliation{Department of Computer Sciences, University of Wisconsin-Madison, Madison, WI 53706, United States}
\alsoaffiliation{Morgridge Institute for Research, Madison, WI 53715, United States}

\author{Spencer S. Ericksen}
\affiliation{Drug Development Core, Small Molecule Screening Facility,
University of Wisconsin Carbone Cancer Center, University of Wisconsin-Madison, Madison, WI 53705, United States}

\author{Anthony Gitter}
\affiliation{Department of Biostatistics and Medical Informatics, University of Wisconsin-Madison, Madison, WI 53715, United States}
\alsoaffiliation{Department of Computer Sciences, University of Wisconsin-Madison, Madison, WI 53706, United States}
\alsoaffiliation{Morgridge Institute for Research, Madison, WI 53715, United States}
\email{gitter@biostat.wisc.edu}

\title[]
  {Chemical Language Model Linker: blending text and molecules with modular adapters}

\begin{document}

%\begin{tocentry}

% Some journals require a graphical entry for the Table of Contents.
% This should be laid out ``print ready'' so that the sizing of the

% Inside the \texttt{tocentry} environment, the font used is Helvetica
% 8\,pt, as required by \emph{Journal of the American Chemical

% The surrounding frame is 9\,cm by 3.5\,cm, which is the maximum
% permitted for  \emph{Journal of the American Chemical Society}
% graphical table of content entries. The box will not resize if the
% content is too big: instead it will overflow the edge of the box.

% This box and the associated title will always be printed on a
% separate page at the end of the document.
%\centering

%\includegraphics[width=9cm,height=3.5cm]{TOC.png}

%\end{tocentry}

\begin{abstract}
The development of large language models and multi-modal models has enabled the appealing idea of generating novel molecules from text descriptions. Generative modeling would shift the paradigm from relying on large-scale chemical screening to find molecules with desired properties to directly generating those molecules. However, multi-modal models combining text and molecules are often trained from scratch, without leveraging existing high-quality pretrained models. Training from scratch consumes more computational resources and prohibits model scaling. In contrast, we propose a lightweight adapter-based strategy named \textbf{Chem}ical \textbf{L}anguage \textbf{M}odel \textbf{L}inker (ChemLML).
ChemLML blends the two single domain models and obtains conditional molecular generation from text descriptions while still operating in the specialized embedding spaces of the molecular domain.
ChemLML can tailor diverse pretrained text models for molecule generation by training relatively few adapter parameters.
We find that the choice of molecular representation used within ChemLML, SMILES versus SELFIES, has a strong influence on conditional molecular generation performance. SMILES is often preferable despite not guaranteeing valid molecules.
We raise issues in using the entire PubChem dataset of molecules and their associated descriptions for evaluating molecule generation and provide a filtered version of the dataset as a generation test set. To demonstrate how ChemLML could be used in practice, we generate candidate protein inhibitors and use docking to assess their quality and also generate candidate membrane permeable molecules.
\end{abstract}

\section{Introduction}
Machine learning methods have emerged as powerful tools in biology and chemistry \citep{ching2018opportunities}. These methods have been  applied extensively across a range of small molecule and protein problems, from molecular property prediction \citep{gilmer2017neural} and molecule interaction prediction \citep{deng2020multimodal} to protein structure prediction \citep{krishna2024RoseTTAFoldAllAtom} and inverse folding design \citep{zheng2023structure}.

Generative machine learning models are a particularly attractive area of research as an alternative to traditional experimental and computational chemical screening.
Directly generating molecules with desired properties can avoid performing high-throughput chemical screens and large-scale virtual screens on molecule databases \citep{alnammi2023evaluating}. By learning the underlying chemical patterns and structures from vast datasets, generative models such as generative adversarial networks \citep{kadurin2017drugan}, variational autoencoders \citep{jin2018junction}, and autoregressive models \citep{bagal2021molgpt} have enabled the automated design of molecules with desired properties.
Recent efforts in generative chemical modeling include controllable cross-modality generation. These methods include contrastive learning, multi-task training, and finetuning of large language models (LLMs).

LLMs have dramatically reshaped the landscape of natural language processing (NLP), demonstrating remarkable capabilities in generating coherent and contextually appropriate text across diverse domains. Built on transformer architectures and trained on broad corpora, LLMs achieve a nuanced representation of language patterns, syntax, and semantics. This advanced representation enables them to perform a wide array of tasks, from simple text completion to complex question answering and summarizing. A critical aspect of their development has been the observed scaling laws \citep{kaplan2020scaling}, which suggest that the performance of these models improves predictably with an increase in model size and data volume. With LLMs' achievements across different domains, researchers are increasingly exploring their potential to address challenges in the fields of biology and chemistry \citep{zhang2024scientific}.

Although general purpose LLMs show some capability to operate with molecules, it is limited \citep{christofidellis2023unifying, microsoft2023impact, mirza2024superhuman}.
A natural question is how to construct a multi-modal model that operates on natural language and chemicals. % Another related paper https://doi.org/10.1101/2023.11.28.568966
Previous strategies include aligning the embedding spaces between text and molecules through contrastive learning, T5-based \citep{raffel2020exploring} text-to-molecule translation, training a single model on multiple molecule-related tasks, and finetuning LLMs on molecule-specific corpora. However, all these methods have varying limitations.
%The initial contrastive learning approach could edit molecules but not generate molecules from scratch.
The T5 model requires additional pretraining and cannot apply previously pretrained models on related areas. 
%For general LLMs, most of the pretraining corpus is irrelevant to the molecular domain. Correspondingly, there will be only a small portion of parameters that are relevant to molecule tasks. %Also, finetuned LLMs are slow at inference.
Most notably, prior work typically combines a single natural language model with a single molecular model without exploring the interactions between models and molecular representations.

To address these issues, we introduce a flexible approach for merging arbitrary text-based language models with molecule generators named \textbf{Chem}ical \textbf{L}anguage \textbf{M}odel \textbf{L}inker (ChemLML). A chemical linker is a molecular structure that connects two or more molecules or functional groups via stable chemical bonds, widely used in drug development (for example, proteolysis targeting chimeras \citep{sakamoto2001protacs} and fragment-based drug design \citep{grenier2023linkers}), biomolecular research, and material science.
% Good PROTAC review https://www.nature.com/articles/s41573-021-00371-6
We borrow the idea of ``linker'', and our ``linker'' metaphorically connects natural and molecular languages through adapters instead of linking different molecules or functional groups.
Despite the name, ChemLML is not specifically designed to generate linker molecules \citep{imrie2020deep, guan2023linkernet, igashov2024equivariant}.
% Also https://doi.org/10.1021/acs.jcim.2c00982, https://doi.org/10.1039/D0SC03126G, https://arxiv.org/abs/2205.07309, https://doi.org/10.1021/acs.jcim.2c01287, https://doi.org/10.1021/acs.jcim.3c01700, etc.
ChemLML makes full use of pretrained models for text and molecules, which can greatly reduce the training time. The main contributions are:
\begin{enumerate}
    \item We present ChemLML, which requires far fewer trainable parameters compared to models of the same scale for text-guided molecule design tasks, achieving strong performance by training adapters only.
    \item ChemLML can flexibly combine multiple types of pretrained text and molecule architectures.
    \item Our case studies suggest that ChemLML may be able to generate candidate inhibitors of protein targets based on its docking performance and membrane permeable molecules from text descriptions alone.
\end{enumerate}

\section{Related Work}
\textbf{Contrastive learning}
Contrastive learning has emerged as a powerful technique for learning rich, meaningful representations in multiple domains.
SimCLR \citep{chen2020simple} performs contrastive learning between different augmented views of the same image to learn visual representations. 
MolCLR \citep{wang2022molecular} extends a similar method to graph neural networks on molecule graphs.
CLIP \citep{radford2021learning} introduces an approach to contrastive learning that unifies language and vision modalities. The model is trained to predict which images are paired with which texts.
MoleculeSTM \citep{liu2023multi} trains a multi-modal molecule-text model by learning a joint representation of molecules’ chemical structures and textual descriptions via contrastive learning. MoleculeSTM is able to perform structure-text retrieval and molecule editing tasks.
CLAMP \citep{seidl_enhancing_2023} utilizes contrastive learning to associate molecules and their corresponding assay descriptions, thus enhancing the model's capability in few-shot and zero-shot molecule activity prediction tasks.

\textbf{Multi-task learning}
Multi-task learning is applied to molecular prediction tasks because it enables models to leverage shared information across tasks, often improving generalization and performance by learning common patterns.
The origins of some multi-modal text-and-chemical models came from NLP.
For instance, MT-DNN \citep{liu2019multi} leverages the Bidirectional Encoder Representations from Transformers (BERT) \citep{devlin2018bert} architecture as a shared encoder and simultaneously learns across multiple NLP tasks. By adding task-specific layers on top of BERT, MT-DNN integrates the advantages of shared knowledge to improve generalization and performance on those tasks.
The T5 model \citep{raffel2020exploring} treats all NLP tasks as text-to-text problems, enabling unified, scalable multi-task learning with a single model. The model is pretrained and finetuned on diverse tasks for enhanced performance and efficiency. MolT5 \citep{edwards2022translation} uses a T5 model to perform molecule-to-text and text-to-molecule tasks at the same time. Text+ChemT5 \citep{christofidellis2023unifying} further extends the number of molecule-related tasks.
%Mol-Instruction \citep{fang2023mol} collects several molecule-oriented tasks. It uses prompts for different tasks to perform instruction tuning on the multi-task dataset.

\textbf{Chemical language models} 
Chemical language models are models that are trained on string representations of molecules \citep{grisoni2023clm}.
They primarily fall into two categories of architectures: BERT and generative pretrained transformer (GPT). ChemBERTa \citep{chithrananda2020chemberta} is a specialized adaptation of the BERT model, tailored for chemistry. It is pretrained with masked language modeling and finetuned for molecule property prediction. MolGPT is a GPT model tailored for molecular generation tasks that is trained by predicting the next token. For a more comprehensive understanding of specific models, see recent reviews \citep{zhang2024scientific, liao2024words, ramos2024review}.
% Chemlactica and Chemma fine tune Galactica and Gemma using computed properties from PubChem molecules that are embedded as text (e.g. QED) https://arxiv.org/abs/2407.18897
% More models https://github.com/hicai-zju/scientific-llm-survey
% Focus on the models we use and extend because there are so many

\textbf{Text-guided molecule generation}
Text2Mol \citep{edwards2021text2mol} proposes the problem of generating molecules from their text descriptions, which is framed as a molecule retrieval task.
MolT5 \citep{edwards2022translation} and 
Text+Chem T5 \citep{christofidellis2023unifying} mentioned previously are examples of models that can generate molecules given text inputs.
nach0 \citep{livne2024nach0} is a multi-modal model also based on the T5 architecture that is pretrained on PubMed abstracts, patent descriptions, and ZINC chemicals.
MoMu \citep{su2022molecular} uses contrastive learning to bridge a graph-based molecule encoder and a natural language encoder to perform multiple molecular tasks.
TGM-DLM \citep{gong_text-guided_2024} leverages diffusion models to address the limitations of autoregressive methods in molecule generation.
MolXPT \citep{liu2023molxpt} includes wrapped sentences in its pretraining dataset, which are sentences from scientific text that contain molecules that are detected and replaced with the matching SMILES string.
MolFM \citep{luo2023molfm} includes a knowledge graph over drugs, proteins, and diseases in its multi-modal model.
AMAN \citep{zhao2024adversarial} introduces a graph transformer for the molecule representation and trains using a combination of adversarial loss and a triplet loss that includes negative samples.
ChemDFM \citep{zhao_chemdfm_2024} performs chemistry domain pretraining and instruction tuning on multiple molecule-related tasks with Llama2, developing an dialogue-based chemical LLM.
Mol-Instruction \citep{fang2023mol} prepares datasets for instruction tuning LLMs, which includes text-guided molecule generation instructions.
MolReGPT \citep{li2024empowering} uses in-context learning with ChatGPT and Llama 2.
In-context learning with ChatGPT can also generate synthetic training data for molecule generation \citep{chen2024artificially}.
MoleculeSTM \citep{liu2023multi} can perform molecule captioning, text-based molecule generation, and text-based molecule editing.
% Instruction-Based Molecular Graph Generation with Unified Text-Graph Diffusion Model https://arxiv.org/pdf/2408.09896
The recent Language + Molecules Workshop \citep{edwards2024lm24}, whose dataset includes biomedical and other chemical properties, also inspired new methods.
A comprehensive review presents additional approaches \cite{pei2024leveraging}.

\section{Methods}
\subsection{Task definition}
We define the task as generating molecules based on a text description such that the molecules match the description. The text may contain information about the desired physical, functional, or chemical properties of the molecule.
\subsection{Preliminaries}
\textbf{Simplified molecular input line-entry system (SMILES)}
SMILES is a notation used to represent molecules as strings \citep{krenn2020SELFIES}. SMILES strings describe atomic composition, connectivity, and stereochemistry using simple rules. For example, the SMILES of 4-methylphenol is Cc1ccc(O)cc1.

\textbf{Self-referencing embedded strings (SELFIES)}
SELFIES is an alternative representation for molecules designed to overcome some limitations of SMILES \citep{krenn2020SELFIES}. It ensures the generation of syntactically and semantically valid molecular structures. This is achieved through a self-referencing system that encodes rules and constraints within the string itself, making SELFIES particularly useful for applications in generative models. The SELFIES of 4-methylphenol is [C][C][=C][C][=C][Branch1][Branch1][C][=C][Ring1][=Branch1][O]. It is much longer than the corresponding SMILES format.

\subsection{ChemLML model architecture}
\textbf{Text pretrained models}
We used three pretrained text LLMs that are specialized for scientific text.
SciBERT \citep{beltagy2019scibert} is a lightweight model with 110M parameters.
Galactica \citep{taylor2022galactica} was reported to have the best performance in a chemistry-related evaluation \citep{mirza2024superhuman}. Although \citet{chartier-edwards_galacticas_2024} pointed out hallucination and performance issues with Galactica, we do not use Galactica's generative capabilities. We use it as an embedding model and explore whether LLMs offer improved scientific text representation capabilities over time in comparison to older models like SciBERT. Due to GPU memory constraints, we only used Galactica-125M, 1.3B, and 6.7B. Finally, we also used the T5 encoder from Text+ChemT5 \citep{christofidellis2023unifying}.

\textbf{Molecule pretrained models} 
We focused on pretrained models that use SMILES or SELFIES string representations. 
MolGPT \citep{bagal2021molgpt} is a GPT model. MolGen \citep{fang2023domain} is a BART-style model that has two stages of pretraining. During the first stage, it randomly masks some of the SELFIES tokens, encodes the corrupted SELFIES using a bidirectional model, calculates the likelihood of the SELFIES string with a left-to-right autoregressive decoder, and calculates the reconstruction loss. In the second stage, it introduces the domain-agnostic molecular prefix as a domain instructor to facilitate the transfer of knowledge across diverse domains. We only use the MolGen autoregressive decoder. MolXPT \citep{liu2023molxpt} is another GPT-style model, which unifies text and molecules during pretraining. It replaces the molecule name with a SMILES string. We only use the molecule decoding capability of MolXPT by feeding the start-of-molecule token at the beginning during decoding.

\begin{figure}[htb]
    \centering
    \includegraphics[width=0.8\textwidth,height=0.8\textheight,keepaspectratio]{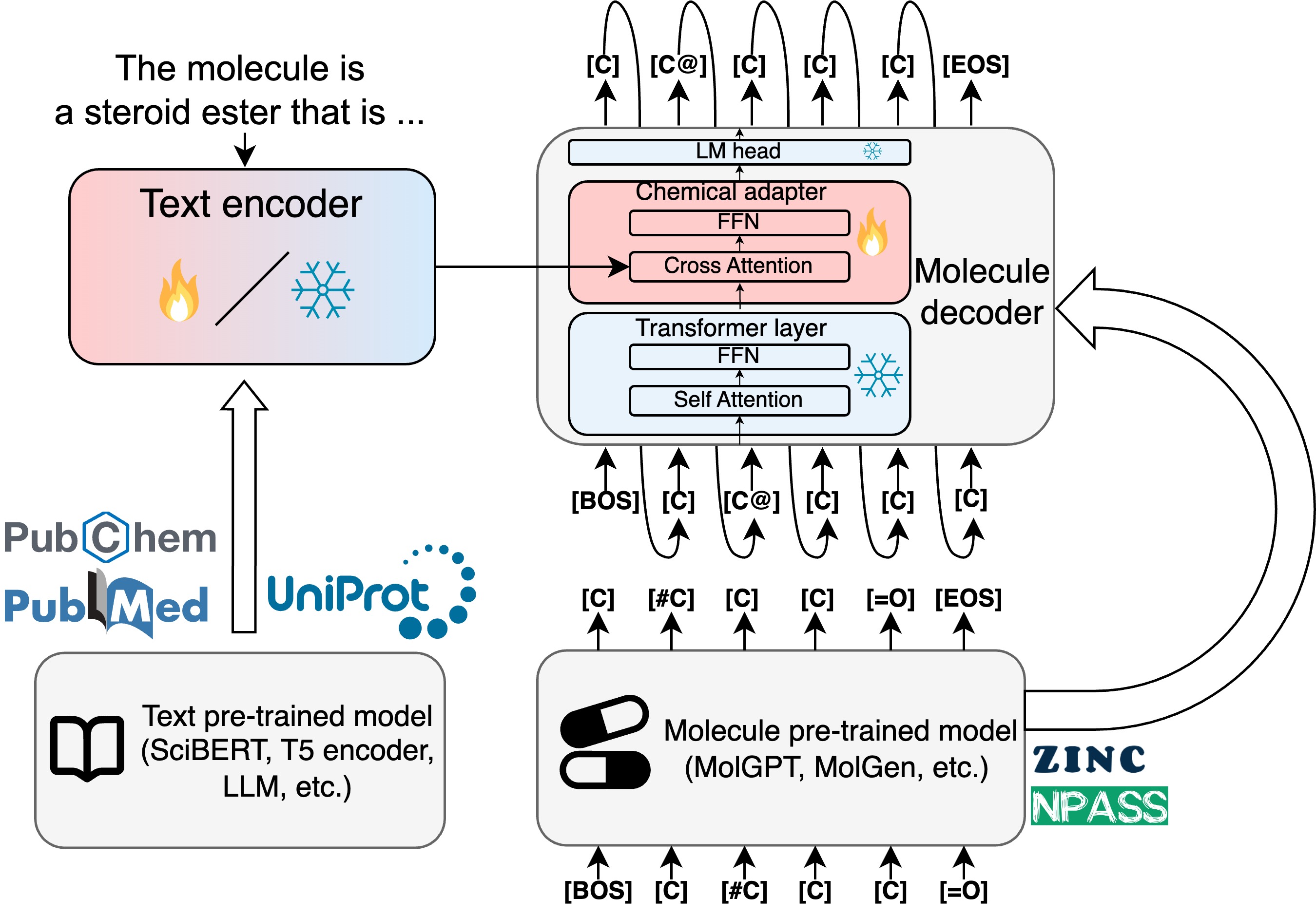}
    \caption{The ChemLML model architecture. We use a pretrained text model as the text encoder and a  pretrained molecule model as the molecule decoder. A single modular chemical adapter is added to the last layer of the molecule decoder. It takes the text and molecule embeddings as input and produces a refined molecule embedding under the guidance of the text embedding. The fire and snowflake icons indicate the parameters are trainable or frozen. FFN: feed-forward network, LM: language modeling, BOS: beginning of sequence, EOS: end of sequence.}
    \label{fig:model}
\end{figure}

\textbf{Complete architecture} We use the pretrained text model as a text encoder and the pretrained molecule model as a molecule decoder. We add a chemical adapter to the last layer of the molecule decoder that takes both text and molecule embeddings as input and outputs a molecule embedding. A similar architecture was applied in the protein domain with LM-Design \citep{zheng2023structure} and ProtT3 \citep{liu_prott3_2024}. The model architecture is shown in Figure \ref{fig:model}.

\subsection{ChemLML training and inference}
\textbf{Adapter module} We use a cross-attention mechanism to construct the adapter between the text embedding and molecule embedding. Let $T=\{t_1, t_2, ..., t_m\}$ be the set of text embeddings and $S=\{s_1, s_2, ..., s_n\}$ be the set of molecule token embeddings. In cross-attention, the text embedding will be projected to the same dimension as the molecule embedding, denoted as $T'=\{t'_1, t'_2, ..., t'_m\}$. We can obtain the cross-attention matrix $a$ with size $m \times n$. For $i=1,2,...,m$ and $j=1,2,...,n$, the attention weights can be calculated as $a_{i,j}=\frac{\exp(t_i^\top s_j')}{\sum_{k=1}^m \exp(t_i^\top s_k')}$ and the molecule embedding will be updated as $s_i'=\sum_{j=1}^{m}a_{i,j}s_j$.
We input the new molecule embedding into the language model head of the molecule pretrained model. The updated embeddings still reside within the embedding space of the original pretrained molecule model. The difference is that these molecules are now aligned with the text embeddings, thereby inheriting the properties described by the text.

During the training process, we use textual descriptions as input, which are first fed into the text LLM model to obtain their embeddings. Subsequently, we use the adapter to perform the cross-attention operation between the text embeddings and molecule embeddings from the last layer of the molecule pretrained model to capture the relationship between the textual and molecular information. We adopt the teacher-forcing strategy during training, using the ground truth token as the input for next step rather than using the model's own output from the last step.

During the inference stage, we employ an autoregressive approach to generate molecules. Specifically, the model generates the molecule sequence in a step-by-step manner, where each step predicts the next sequence token based on the previously generated sequence and the text embedding. This process can be expressed as:
\begin{equation}
    y_t \sim P(y | y_{<t}, T; \theta)
\end{equation}
where $y_t$ is the token generated at time step $t$, $y_{<t}$ denotes the sequence generated prior to time step $t$, $T$ is the given text embedding, and $\theta$ represents the model parameters.

\subsection{Metrics}
We intend generated molecules to exhibit specific properties.
Following the molecular similarity principle \citep{duranfrigola2020similarity}, this would result in the generated molecules having structures similar to the ground truth molecules, thereby achieving a high degree of similarity and low diversity during evaluations.
Therefore, we primarily focus on various similarity metrics and calculate the similarity between generated molecules and ground truth molecules. We use MACCS fingerprints, RDKit fingerprints \citep{rdkit}, and Morgan fingerprints as different molecule representations that produce different similarity scores. The similarities between these fingerprints are calculated with Tanimoto similarity. We discuss other metrics not used here in the Supplementary Methods.

SMILES-based generation sometimes generates syntactically invalid molecules, so we also include validity as a metric. Although valid SELFIES strings are guaranteed to be syntactically valid chemicals, the SELFIES-based model does not always generate valid SELFIES strings, such as output strings with an [EOS] token at the beginning. The validity metric only evaluates syntactic correctness without accounting for the molecule's 3D structure.

\subsection{Datasets}
\textbf{Pretraining datasets}
The pretrained molecule generation models were trained on the following datasets:

\begin{itemize}
    \item Selected ZINC-15: Chemformer \citep{irwin2022chemformer} selected approximately 100M molecules from 1.5B available molecules from ZINC-15 \citep{sterling2015zinc} with the following constraints: reactivity set to reactive, purchasability set to annotated, molecular weight $\leq 500$ Daltons, and LogP $\leq 5$. MolGen uses this dataset during the first stage of pretraining.
    \item NPASS: Natural Product Activity \& Species Source Database (NPASS) \citep{zeng2018npass} is a natural product domains database, which is used in the second pretraining stage of MolGen.
    \item MOSES: MOSES \citep{polykovskiy2020molecular} is a cleaned version of ZINC, containing 1.9M molecules. It removes the atoms besides C, N, S, O, F, Cl, Br, and H and cycles longer than 8 atoms. It does not contain chirality information. Models pretrained on the MOSES dataset, such as MolGPT, are unable to process SMILES strings that include atoms beyond this dataset's scope or SMILES strings that contain chirality information.
    \item PubChem and PubMed: MolXPT trains on the titles and abstracts from 30M PubMed entries, SMILES strings from 30M chemicals from PubChem \citep{kim2023pubchem}, and 8M sequences in which chemical names are replaces with SMILES.
\end{itemize}

We do not train directly on these datasets.
However, information about the training data of the molecule generation models that ChemLML builds upon is helpful for understanding its molecule generation performance.

\textbf{Molecule description datasets} We use two different molecule description datasets.
We train and test all the models on ChEBI-20 and use PubChem as an additional test set. We use the ChEBI-20 training set for training. Then, we tune the hyperparameters based on the convergence status on the ChEBI-20 training set and the performance on the ChEBI-20 validation set. Finally, we test on the ChEBI-20 and PubChem test sets.

\begin{itemize}
    \item ChEBI-20: The Chemical Entities of Biological Interest (ChEBI) database \citep{hastings2016chebi} is a dictionary of small molecules paired with text descriptions.
    \citet{edwards2021text2mol} created the ChEBI-20 dataset by combining PubChem compounds with ChEBI annotations and retaining the 33,010 compound-description pairs for which the description had more than 20 words.
    % Sources https://www.ebi.ac.uk/chebi/aboutChebiForward.do (no PubChem)
    % PubChem lists ChEBI as a source though https://pubchem.ncbi.nlm.nih.gov/source/ChEBI
    % See all 991 sources https://pubchem.ncbi.nlm.nih.gov/sources
    The ChEBI-20 training, validation, and test set splits of 80\%, 10\%, and 10\% are also from \citet{edwards2021text2mol}.
    
    \item PubChem: In our evaluations, we assume that each text description is informative and associated with one unique molecule. Many examples in the complete PubChem dataset \citep{kim2023pubchem} do not meet that assumption, so we apply several types of filters on the molecules and text descriptions. In short, we select descriptions greater than 30 words and remove one-to-many description-molecule pairs. In this way, we remove ambiguous instances and improve the quality of our evaluations.
    After filtering, we remove the intersection of molecules between PubChem and ChEBI-20 dataset and obtain a dataset that has 11,563 examples. We randomly sample 3,000 examples from this dataset, which we refer to as PubChem-filtered, and use it as an additional test set for models trained on ChEBI-20.
    %\ Why are we not using all of them?
    To demonstrate the impact of this filtering, we also report results on 2,576 examples sampled from the unfiltered PubChem dataset, which we refer to as PubChem-unfiltered. More details of PubChem dataset filtering and processing are discussed in Supplementary Methods.
\end{itemize}

\subsection{Baseline methods}
\textbf{T5}
The T5 model comes from the MolT5 paper \citep{edwards2022translation}. It is trained on the ChEBI-20 dataset.

\textbf{MolT5}
MolT5 loads the T5.1.1 checkpoint and is pretrained using the replace corrupted spans objective. During each pretraining step, a minibatch comprising both natural language sequences and SMILES sequences is sampled and certain words are randomly selected for corruption. The task is to predict the spans of tokens that were corrupted. In the finetuning stage, the model is trained on ChEBI-20.

\textbf{MolXPT} MolXPT is pretrained on PubMed and PubChem data. Then, it is finetuned on ChEBI-20 dataset.

\textbf{TGM-DLM} TGM-DLM \citep{gong_text-guided_2024} uses the text embedding to guide a diffusion language model to generate molecules. It employs a two-phase diffusion process, first generating a molecule based on the text description followed by correcting invalid SMILES strings.

\textbf{Text+ChemT5} Text+ChemT5 is a multi-task, multi-domain model for natural and chemical language. It is trained on the four tasks listed in the Supplementary Methods.

\textbf{T5 encoder+MolXPT MLP} We introduce this baseline model as an ablation for ChemLML to demonstrate the value of the full ChemLML model over its individual components.
This model concatenates the T5 text and MolXPT molecule embeddings and passes them through a two-layer multilayer perceptron (MLP).

\subsection{Training and evaluation} 
We evaluate the models' performance on text-based molecule generation. For MolT5, Text+ChemT5, MolXPT, SciBERT, and different scales of MolGen, we use their checkpoints on HuggingFace.  For the Text+ChemT5 encoder, we add the prompt ``Write in SMILES the described molecule'', which is used in the corresponding task in Text+ChemT5.

For ChemLML, we use SciBERT, the encoder from Text+ChemT5, and different scales of Galactica as the text encoders and combine them with MolGPT, MolXPT, MolGen, and MolGen-7B as the molecule decoders. The weights of the molecule decoder are always frozen. The weights of the adapter are trainable. The weights of the text encoder can be either trainable or frozen.  We run experiments on two datasets, ChEBI-20 and PubChem.

For ChEBI-20 baseline evaluations, we obtain the results from the MolT5, Text+ChemT5, MolXPT, and TGM-DLM publications. The T5 results are also obtained from the MolT5 publication.

MolGPT can not be finetuned on ChEBI-20 because it is pretrained on the MOSES dataset so its tokenizer does not interpret chirality information. We use MolGPT to compare SMILES- and SELFIES-based models. First, we use RDKit to remove stereochemical information from all of the molecules. Then, we remove the molecules that fail to be tokenized by MolGPT's tokenizer. Lastly, we transform all the SMILES to SELFIES. This reduces the ChEBI-20 training set from 26,407 to 15,899 instances, validation set from 3,301 to 1,961 instances, and test set from 3,300 to 2,032 instances. We build the MolGPT model with 12 layers, 8 attention heads, and a 256-dimensional embedding. The only differences between the MolGPT SMILES and SELFIES models are the output dimensions, depending on the vocabulary lists of each method.

We use the Noam Optimizer with 4,000 warm-up steps for training. For molecule sampling, we use multinomial sampling with a random seed of 42 for all methods unless noted otherwise.

\subsection{Docking case study setting}
We filtered the PubChem dataset to identify text descriptions that pertain to molecules that inhibit specific protein targets by initially retaining only descriptions with the string ``inhibit''.
Then, we split all remaining instances into low (0.15 to 0.3), medium (0.3 to 0.5), and high (0.5 to 0.9) similarity based on the Morgan fingerprint similarity between the ground truth and generated molecule from the ChemLML(T5 encoder+MolGen) model.
We selected multiple examples per similarity bin (Table~\ref{tab:docking}) as described in the Supplementary Methods.

After we generated the first example molecule with the temperature set to 1 and random seed 42, we set the temperature to 1.5 and changed the random seed in order to sample different molecules. However, in some cases, the conditions specified in the conditional molecule generation task were overly restrictive. The model generated only a few unique molecules even when sampling 1,000 times. In these cases, we iteratively increased the temperature to encourage diversity and obtain the desired number of unique molecules. We generated up to 1,000 molecules per temperature. If the target number of unique molecules had not yet been obtained, we increased the temperature. The maximum temperature we reached with this procedure was 4.5. The iterative process and variability across prompts made it difficult to track the number of effective sampling iterations required to generate the target number of unique molecules.
As a representative example, we report the number of duplicate samples in the permeability case study (Table \ref{tab:permeability_count}).

We canonicalized the generated SMILES with RDKit and removed duplicates. 
We repeated the iterative generation process above until we obtained 99 additional unique and valid SMILES, resulting in a total of 100 ChemLML-generated molecules per target. 
The SMILES were then used to build 3D conformers with OpenEye's Omega2, and only a subset produced valid 3D structures for docking (Supplementary Methods).
For ChemLML, we used the T5 encoder+MolXPT and T5 encoder+MolGen settings.
We omit ``encoder'' for brevity in the docking experiments. We performed the same molecule generation process with the MolT5 and Text+ChemT5 baselines.

For each of eight target proteins, we docked the ground truth molecule, generated molecules, and control molecules from two types of background distributions: FDA-approved compounds and ChemLML(T5+MolGen)-generated molecules from randomly-sampled descriptions.
A consensus docking score was computed for each molecule that combines results from four docking programs (Supplementary Methods).
We hypothesized that the ground truth molecule and generated molecules resembling this ground truth molecule would score more favorably on their respective protein targets than background molecules. 

\subsection{Permeability case study setting}
We initially used the prompt ``The molecule has high membrane permeability''.
However, we found that MolT5 and Text+ChemT5 generated molecules containing protons such as ``[H+].[H+].[H+].CN(C=O)C=O.CI'' and nonsensical combinations of natural and chemical language like ``CN (C)CI via minimal irritation on minimal water condition.CNC''. Although these can be parsed by RDKit, they do not represent chemically valid or meaningful molecules. Therefore, we modified the prompt to ``The molecule is a drug-like compound with high passive membrane permeability. It contains no formal charge and avoids ionizable groups''. We used a script to remove invalid molecules, elemental substances, salt fragments, and other irrelevant side chains, retaining only the main structure. A generated result is considered valid only if it remains unchanged after this filtering process. We used a temperature of 2.0 and continued sampling by switching the seed until we obtained the desired number of molecules.

\subsection{Data and Software Availability}
The ChemLML code and PubChem-filtered dataset are available from \url{https://github.com/gitter-lab/ChemLML} and archived at \url{https://doi.org/10.5281/zenodo.13925649}.
The pretrained ChemLML models and PubChem-unfiltered dataset are available from \url{https://zenodo.org/doi/10.5281/zenodo.11661517}.
The ChEBI-20 dataset is available from \url{https://github.com/cnedwards/text2mol}, and a copy is in the ChemLML GitHub repository.

\section{Results}

We first compare baseline and ChemLML models trained on ChEBI-20 text descriptions on the ChEBI-20 test set (Table~\ref{tab:CheBI-20 result}). Some versions of the ChemLML models finetune the text encoder and some use the frozen text encoder. The best performing ChemLML model is the T5 encoder finetune+MolXPT version. With 114M trainable parameters, this model achieves 0.727 in Morgan FTS.
It is not surprising to find that models that include the T5 encoder work the best because the T5 encoder has been trained on multiple tasks. The ChemLML combination T5 encoder+MolXPT performs better than the baseline MolT5 even though MolT5 has 52 times more trainable parameters.

\begin{table}[htb]
    \centering
\resizebox{\textwidth}{!}{%
\begin{tabular}{l|cccccccc}
\toprule
&Format &Models& Trainable/Total Params. & Exact $\uparrow$  &MACCS FTS $\uparrow$&RDK FTS $\uparrow$ & Morgan FTS $\uparrow$& Validity $\uparrow$\\
\hline
\multirow{3}{*}{\parbox{2cm}{Baseline methods}}
&SMILES &T5 &247M/247M & 0.069& 0.731& 0.605& 0.545&0.660\\
&SMILES &MolT5 & 248M/248M & 0.081 & 0.721 & 0.588 & 0.529 &  0.772 \\
&SMILES &MolXPT & 350M/350M & 0.215 & 0.859 & 0.757 & 0.667 & \textbf{0.983} \\
&SMILES &TGM-DLM & 180M/180M &\textbf{0.242} & 0.854 & 0.739 & 0.688 & 0.871 \\
&SMILES &Text+ChemT5 & 223M/223M &  0.212 & \textbf{0.874} & \textbf{0.767} & \textbf{0.697} & 0.792 \\
& SMILES &T5 encoder+MolXPT MLP & 111M/461M & 0.227 & 0.820 & 0.713 & 0.636 & 0.980 \\
% T5 & 0.762& 0.069& 24.950& 0.731& 0.605& 0.545&2.48&0.660\\
% MolT5 & 0.769 & 0.081 & 24.458 & 0.721 & 0.588 & 0.529 & 2.18 &  0.772 \\
% Text+ChemT5 & 0.750 & 0.212 & 27.39 & 0.874 & 0.767 & 0.697 & 0.061 & 0.792 \\
\hline
\multirow{11}{*}{{ChemLML}}
&SELFIES &SciBERT+MolGen & 4.7M/317M & 0.083 &  0.680 &0.526&0.422&0.995 \\
& SELFIES &Galactica 125M+MolGen& 4.7M/333M&  0.072& 0.687& 0.533& 0.433&0.994\\
% T5 encoder + MolGen & &0.145 & &0.738 &0.601 & 0.498 & 1.45 & 0.994 \\
&SELFIES &T5 encoder+MolGen & 4.7M/317M& 0.165& 0.758 & 0.616 & 0.527 & 0.995 \\
&SMILES &T5 encoder+MolXPT & 4.7M/464M& \textbf{0.305} & \textbf{0.837} & \textbf{0.745} & \textbf{0.674} & 0.986 \\
& SELFIES &Galactica 1.3B+MolGen& 7.4M/1.53B &  0.091& 0.706& 0.560& 0.454&0.995\\
& SELFIES &Galactica 6.7B+MolGen& 11.5M/6.87B & 0.103& 0.732& 0.588& 0.480&\textbf{0.997}\\
& SELFIES&Galactica 125M+MolGen 7B & 56.7M/6.66B& 0.134 & 0.732& 0.608& 0.522& 0.995\\
&SELFIES &T5 encoder+MolGen 7B & 56.7M/6.64B &0.209 &0.777&0.666 &0.581 & 0.995\\
%\hline
\cline{2-9}
&SELFIES&SciBERT finetune+MolGen & 115M/317M &0.176& 0.757  & 0.633 & 0.547 & \textbf{0.996}\\
&SELFIES&T5 encoder finetune+MolGen & 114M/317M &0.250 & 0.790 & 0.667 & 0.587 & 0.994\\
&SMILES&T5 encoder finetune+MolXPT & 114M/464M & \textbf{0.389} & \textbf{0.868} & \textbf{0.790} & \textbf{0.727} & 0.987 \\
&SELFIES&Galactica 125M finetune+MolGen &130M/333M&  0.143 &  0.743 & 0.606& 0.510& 0.995 \\
&SELFIES&Galactica 125M finetue+MolGen 7B& 182M/6.66B & 0.170& 0.759& 0.640& 0.554&0.995\\
&SELFIES&T5 encoder finetune+MolGen 7B&166M/6.64B& 0.234 &0.793 &0.678 & 0.596 &0.995\\
 \bottomrule
\end{tabular}
}
    \caption{Results of molecule generation on the ChEBI-20 test set. The T5, MolT5, MolXPT, TGM-DLM, and Text+ChemT5 results are copied from their respective papers.  The ChemLML models are grouped into those that finetune the text model and those that do not. We bold the best model per metric within each model category: baseline, ChemLML without finetuning, and ChemLML with finetuning. FTS: Fingerprint Tanimoto similarity }
    \label{tab:CheBI-20 result}
\end{table}

\subsection{Comparison among pretrained models used with ChemLML}

When comparing the ChemLML models that use the MolGen model and do not finetune the text model, it is surprising that the cross-modal representation capability of similarly sized text models does not grow over time. The ChemLML variants SciBERT+MolGen and Galactica 125M+MolGen performed similarly. SciBERT was published in 2018 with 110M parameters and is trained on 3.3B tokens. Galactica 125M is trained on 106B tokens, around 32 times more than SciBERT. A possible explanation is that Galactica 125M is too small for the large training set. However, Galactica 1.3B also outperforms SciBERT by only a small margin. Even Galactica 6.7B, which is far larger than Galactica 125M, only yields a slight improvement. LLMs perform well when combined with MolGen 7B. ChemLML Galactica 125M+MolGen 7B achieves better results than MolT5.

We also consider the results from finetuning the pretrained language models along with the adapter. SciBERT outperforms Galactica 125M in this setting. Also, ChemLML SciBERT finetune+MolGen outperforms MolT5 with half the trainable parameters. The combination of the T5 encoder and MolGen is further strengthened by finetuning the T5 encoder. However, for the ChemLML T5 encoder models that use MolGen 7B, the improvement from finetuning the T5 encoder is marginal.

Based on the values in Table \ref{tab:CheBI-20 result}, it seems that ChemLML T5 encoder+MolGen does not outperform the baseline Text+ChemT5 on all similarity metrics at first glance.
However, the results are influenced by the trade-off between similarity and validity and the selection of SMILES and SELFIES, which we expand upon below.

\subsection{Molecule similarity and validity trade-off}
There is a trade-off between generated and ground truth molecule similarity and the generated molecule validity (Table \ref{tab:CheBI-20 result}). T5 and MolT5 have the same model architecture. MolT5 has been pretrained on a large amount of data with both SMILES and natural language while T5 has not. At first glance, it is therefore surprising that T5 outperforms MolT5 in the similarity metrics. However, MolT5 generates 11 percentage points more valid molecules than T5, which means more molecules are evaluated by the similarity calculation. These molecules that are more difficult to generate may have reduced the mean similarity.

\subsection{SMILES versus SELFIES representations}
The choice of molecule representation, SMILES or SELFIES, is another factor that influences the similarity calculations.
\citet{skinnider2024invalid} recently showed that when training molecule language models on samples from ChEMBL, molecules generated from SMILES-based models matched the training set much better than SELFIES-based models. Invalid SMILES are low-likelihood samples, and filtering low-likelihood outputs improves performance on distribution-learning metrics.
% Figure 1e result from skinnider2024invalid
However, MolGen is trained on SELFIES, and there is not yet an equivalent model pretrained on SMILES that goes through multiple stages of pretraining. 
Therefore, we use MolGPT to compare SMILES and SELFIES directly. 

\begin{table}[htb]
    \centering
\resizebox{\textwidth}{!}{%
\begin{tabular}{cccccccc}
\toprule
Representation &  Trainable/Total Params&Exact $\uparrow$ &MACCS FTS $\uparrow$&RDK FTS $\uparrow$ & Morgan FTS $\uparrow$ & Validity $\uparrow$\\
\hline

\hline
SMILES & 0.59M/120M   &\textbf{0.035}& \textbf{0.704}& \textbf{0.508}& \textbf{0.454}&0.496\\
SELFIES & 0.59M/120M   &0.020 &  0.620 & 0.398 & 0.308 &  \textbf{1.000} \\
SELFIES (on valid SMILES) & 0.59M/120M  &0.021& 0.616 & 0.400 & 0.303 & 1.000\\
\bottomrule
\end{tabular}
}
    \caption{Comparison of MolGPT models trained on SMILES and SELFIES representations.}
    \label{tab:SMILES and SELFIES}
\end{table}

After MolGPT models are pretrained on the MOSES dataset, we train ChemLML T5 encoder+MolGPT models on the ChEBI-20 training dataset and evaluate the performance on the test set (Table~\ref{tab:SMILES and SELFIES}). There is a large difference between the performance of SMILES and SELFIES versions of the same model. When we limit the molecules to the intersection of valid molecules between both methods to ensure that the results are unaffected by the similarity and validity trade-off we mention above, the SELFIES method still performs poorly. The SMILES method outperforms the SELFIES method by about 50\% in the Morgan fingerprint similarity metric. Therefore, the conclusions of the prior study about unconditional molecule generation \citep{skinnider2024invalid} also hold for conditional molecule generation.

Given that SMILES substantially outperform SELFIES in this analysis, we focus on comparing different pretrained models instead of comparing SMILES- and SELFIES-based methods.

\subsection{PubChem evaluation reveals pros and cons of LLMs}
We use molecule descriptions in PubChem as an additional molecule generation evaluation beyond ChEBI-20.
Because many of the PubChem molecule descriptions are overly ambiguous and generic, we prioritize the PubChem-filtered version of the dataset but also evaluate the PubChem-unfiltered dataset to assess how filtering affects the results.
On PubChem-filtered, LLMs without finetuning exhibit limited performance compared with T5 encoders that have been trained on relevant examples. After finetuning, LLM encoders show comparable results (Table~\ref{tab:PubChem_result}).
PubChem-filtered has a similar distribution as the ChEBI-20 dataset, so the ChemLML T5 encoder+MolXPT model still outperforms other methods that have a frozen text encoder.
The ChemLML combination T5 encoder finetune+MolGen 7B does not perform well, potentially due to overfitting on the ChEBI-20 dataset. Another notable finding is that ChemLML generates a higher proportion of exactly matched molecules compared to the MolT5 and Text+ChemT5 baselines. We suggest that because MolGen and MolXPT have been pretrained on a large number of molecules, ChemLML is able to retrieve those molecules accurately based on the text embedding.

\begin{table}[htb]
    \centering
\resizebox{\textwidth}{!}{
\begin{tabular}{l|ccccccc}
\toprule
& Models&Trainable/Total Params& Exact $\uparrow$  &MACCS FTS $\uparrow$&RDK FTS $\uparrow$ & Morgan FTS $\uparrow$ & Validity $\uparrow$\\
\hline
\multirow{2}{*}{\parbox{2cm}{Baseline methods}}
& Text+ChemT5 &223M/223M &  0.028& 0.682 & 0.584 & 0.495 & 0.696 \\
& MolT5 & 248M/248M & 0.052& 0.672 & 0.600 & 0.594& 0.616 \\
\hline
\multirow{6}{*}{{ChemLML}}
& T5 encoder+MolGen &4.7M/317M & 0.165& 0.588 & 0.477 & 0.376 & 0.967 \\
& T5 encoder+MolXPT &4.7M/464M & \textbf{0.247} & \textbf{0.610} & \textbf{0.526} & \textbf{0.428} & 0.978 \\
& Galactica 125M+MolGen& 4.7/333M & 0.002& 0.458& 0.311& 0.204&0.968\\
& Galactica 1.3B+MolGen & 7.4M/1.53B & 0.004& 0.489& 0.332& 0.221&0.958\\
& T5 encoder+MolGen 7B & 56.7M/6.64B &0.004&0.483&0.341&0.225&\textbf{0.988} \\
\cline{2-8}
& T5 encoder finetune+MolGen &114M/317M & 0.215& 0.598& 0.505& 0.404&0.973 \\
& T5 encoder finetune+MolXPT & 114M/464M & \textbf{0.286} & \textbf{0.624} & \textbf{0.543} & \textbf{0.453} & 0.972\\
& Galactica 125M finetune+MolGen &130M/333M & 0.191& 0.566& 0.469& 0.372&\textbf{0.978}\\
& T5 encoder finetune+MolGen 7B & 166M/6.64B& 0.008 & 0.532 & 0.388&0.265 & 0.966\\ 
\bottomrule
\end{tabular}
}
    \caption{Results of molecule generation on the PubChem-filtered test set. SciBERT is excluded because its encoder cannot handle inputs exceeding 512 tokens. The ChemLML models are grouped into those that finetune the text model and those that do not. The highest value in each performance category is shown in bold.}
    \label{tab:PubChem_result}
\end{table}

We are cautious interpreting results from the PubChem-unfiltered dataset (Table~\ref{tab:PubChem_unfiltered_result}) due to the abundance of generic descriptions (Figure~\ref{fig:frequency}), but they do illuminate difference among the ChemLML models.
We observe that the performance of Galactica 125M surpasses both SciBERT and the T5 encoder when they are all used with frozen text encoders and the MolGen model, contrary to what was observed with the ChEBI-20 dataset. 
The large pretraining corpus on scientific texts might account for this. It could enable the models to perform better on unseen descriptions.

\subsection{Docking case study}
\begin{figure}[htb]
    \centering
    \includegraphics[width=1.0\textwidth,height=1.0\textheight,keepaspectratio]{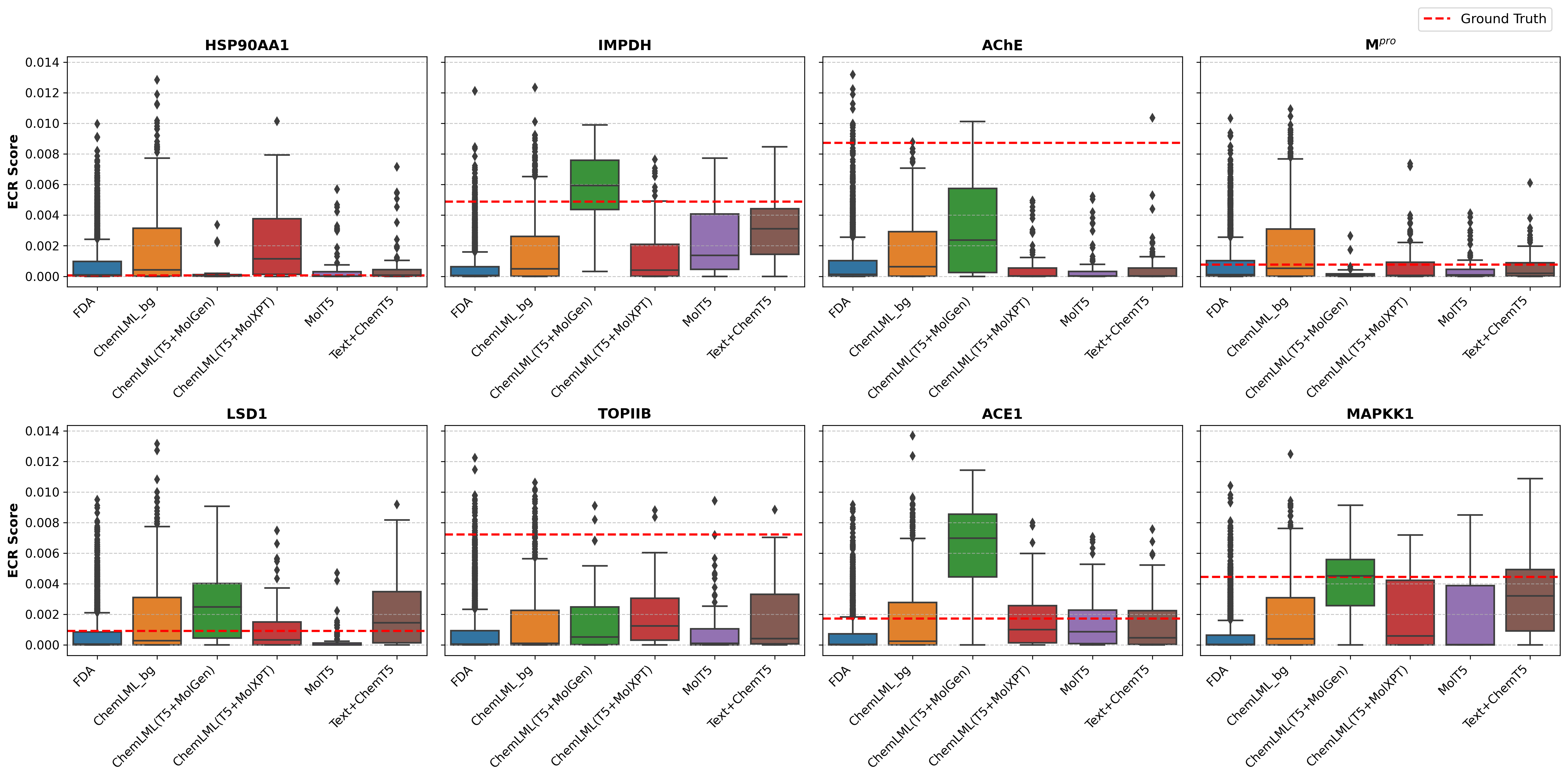}
    \caption{Higher scores on the y-axis are better. Sample sizes for the docked compound sets are displayed in Table \ref{tab:Generated_Cpd_Attrition}. ECR: exponential consensus ranking}
    \label{fig:docking_hist}
\end{figure}

\begin{figure}[htb]
    \centering
\includegraphics[width=\textwidth]{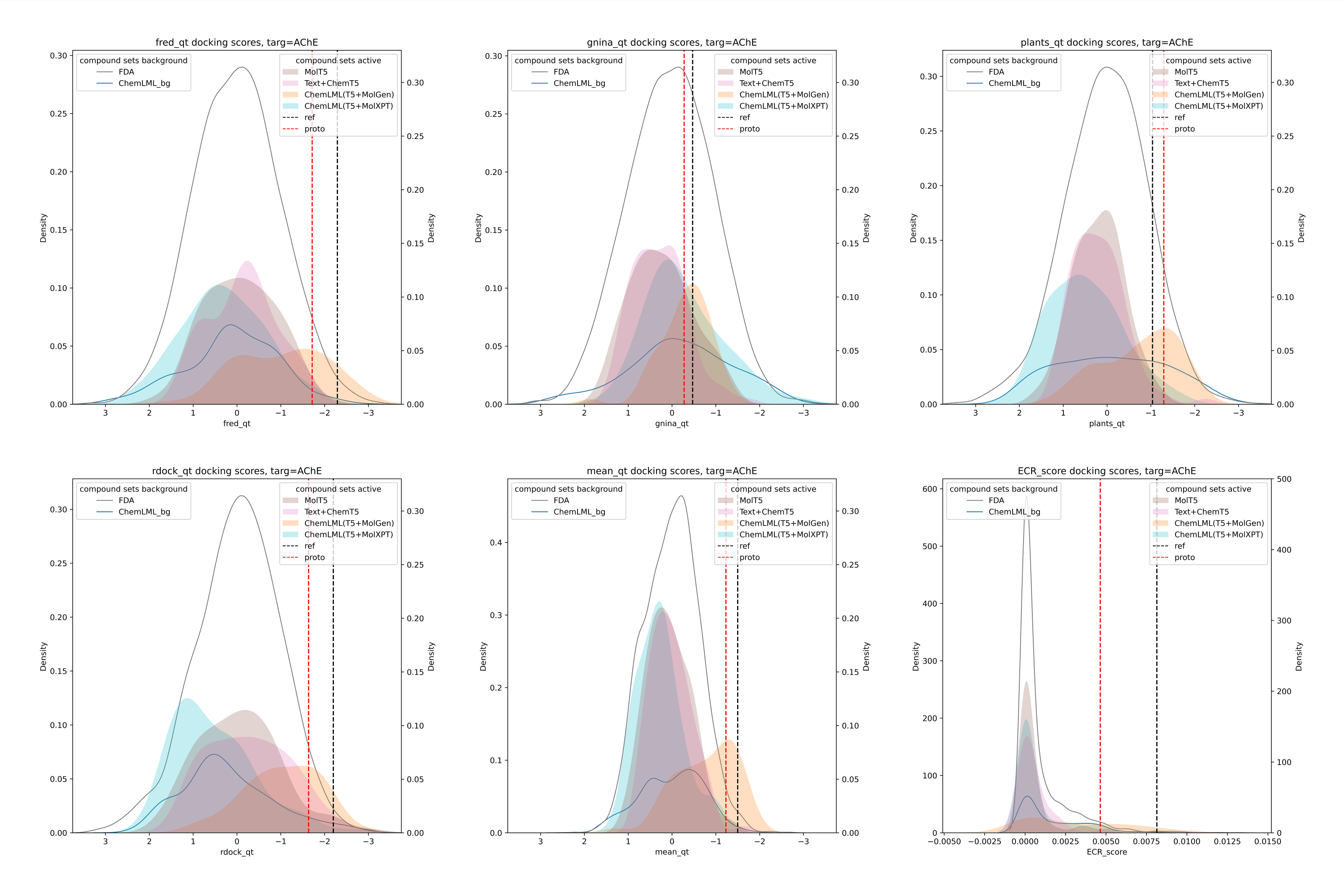}
    \caption{Individual docking program scores for AChE.}
    \label{fig:individual_AChE}
\end{figure}

\begin{figure}[htb]
    \centering
\includegraphics[width=\textwidth]{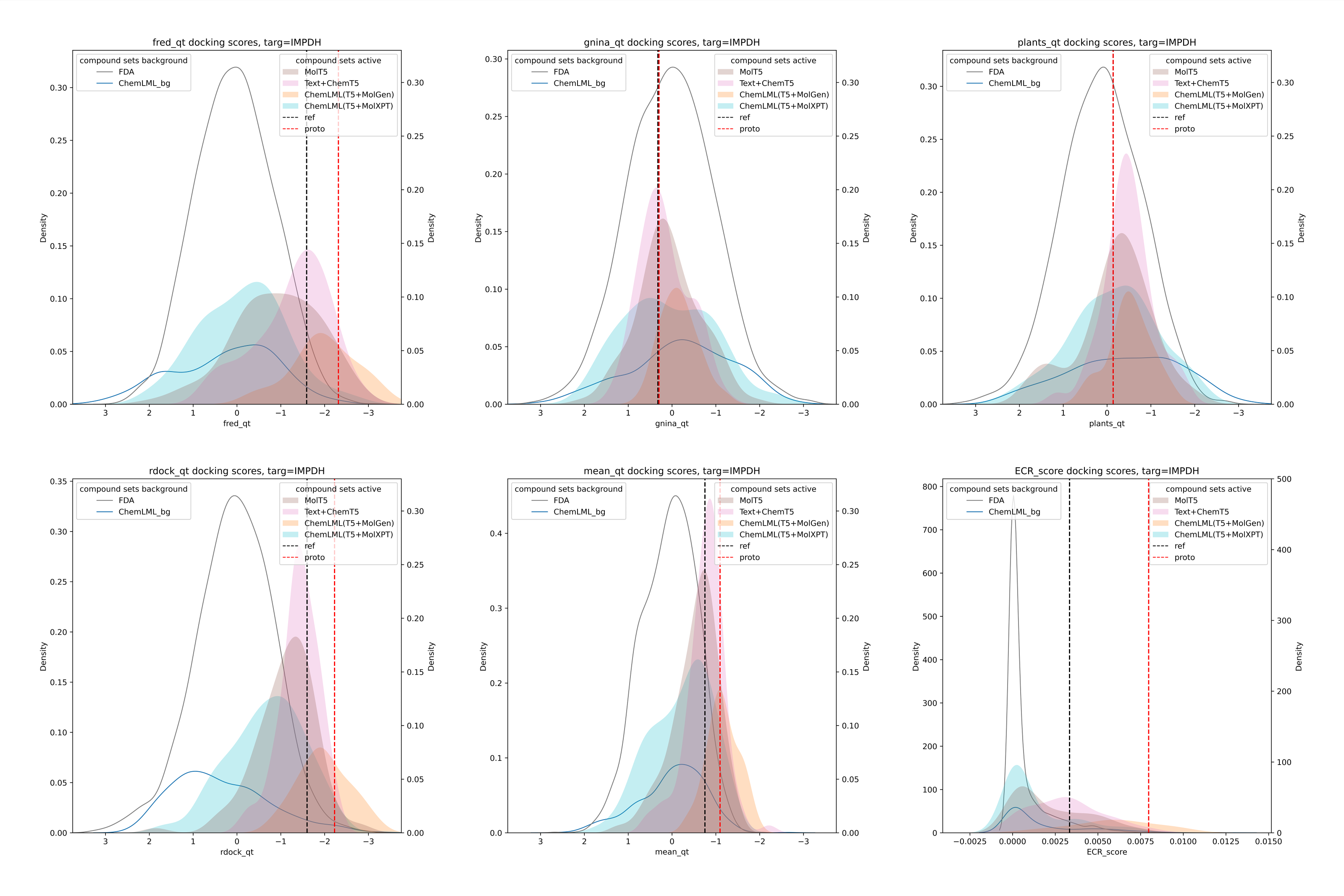}
    \caption{Individual docking program scores for IMPDH.}
    \label{fig:individual_IMPDH}
\end{figure}

\begin{figure}[htb]
    \centering
\includegraphics[width=\textwidth]{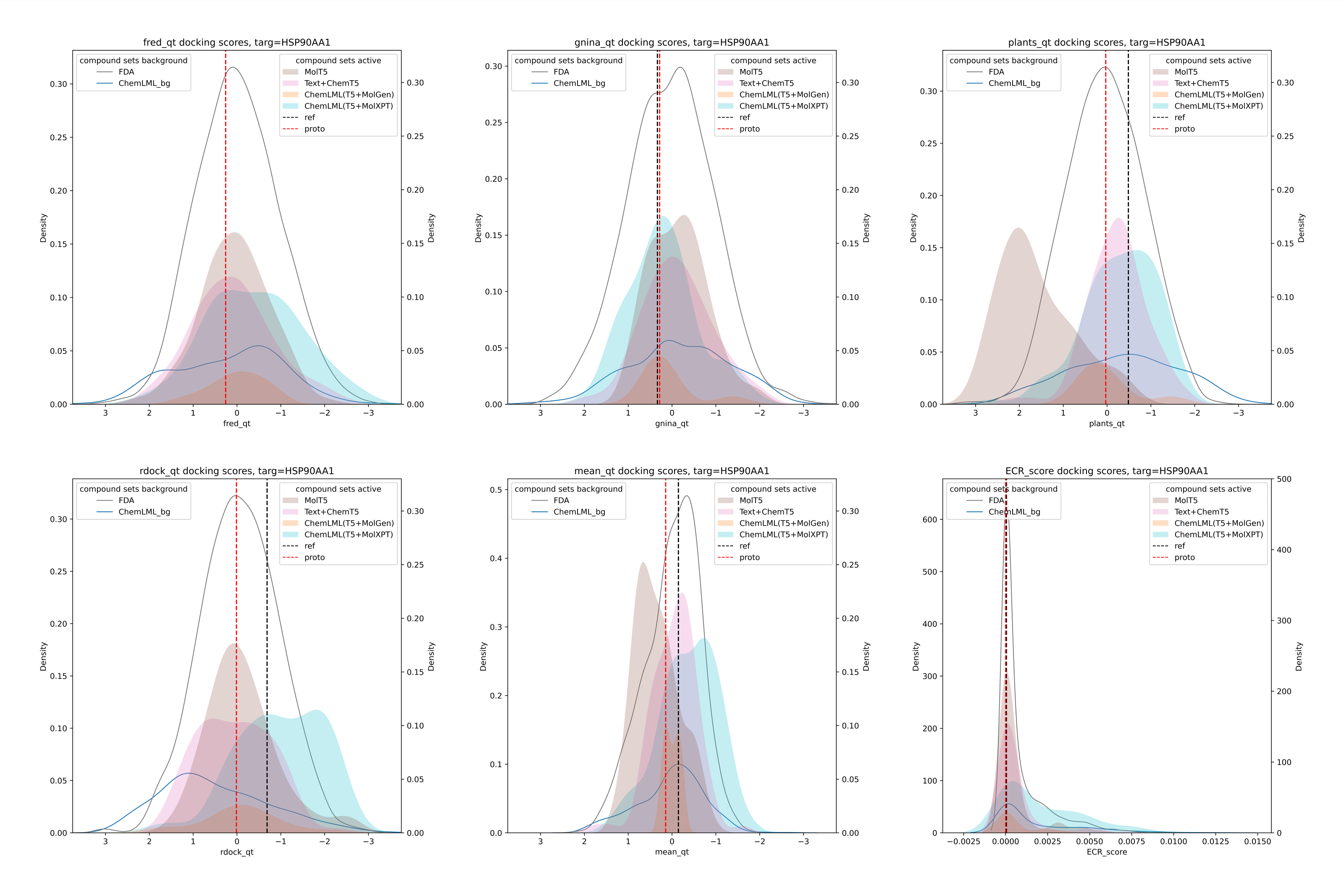}
    \caption{Individual docking program scores for HSP90AA1.}
    \label{fig:individual_HSP90AA1}
\end{figure}

\begin{figure}[htb]
    \centering
\includegraphics[width=\textwidth]{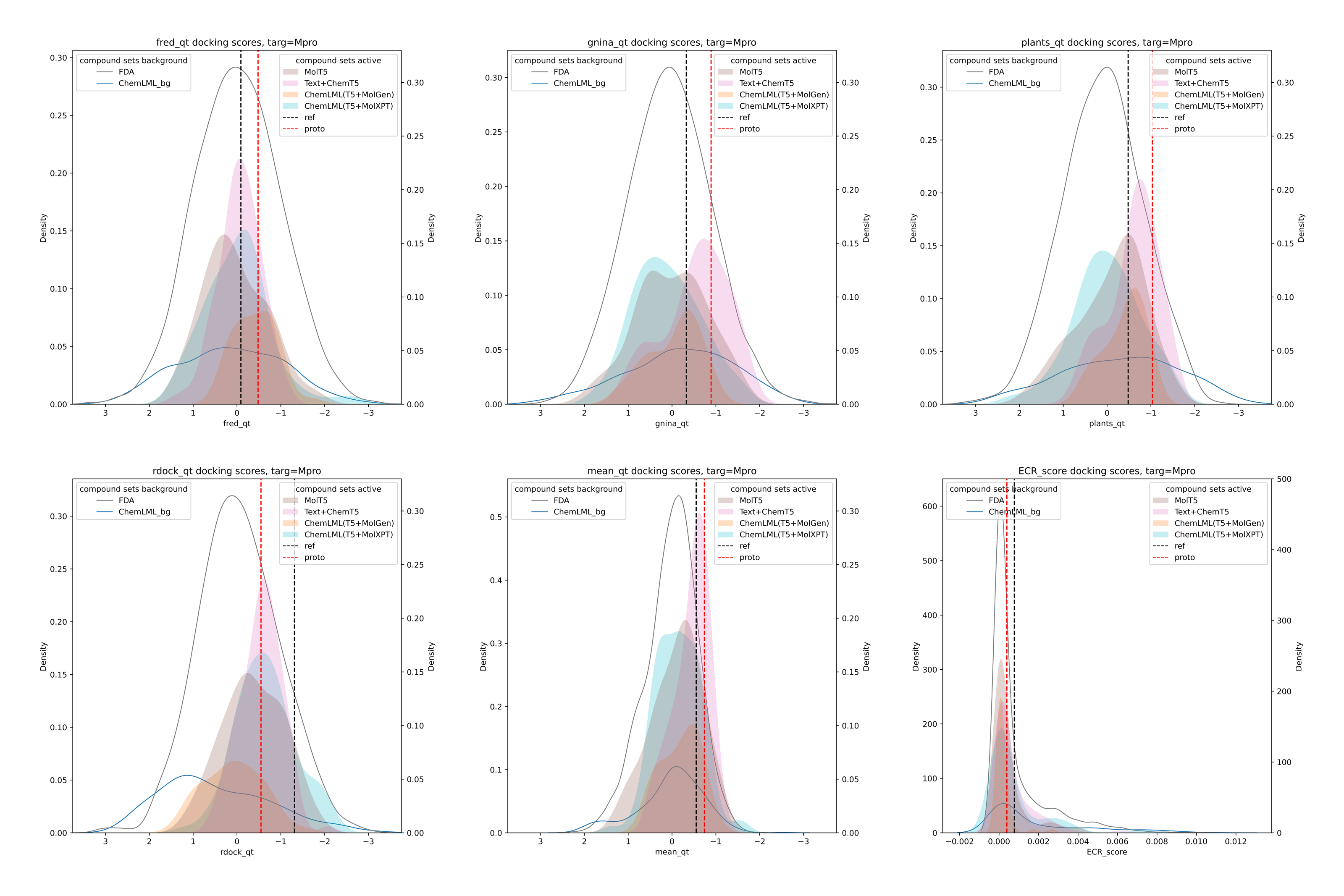}
    \caption{Individual docking program scores for M\textsuperscript{pro}.}
    \label{fig:individual_Mpro}
\end{figure}

\begin{figure}[htb]
    \centering
\includegraphics[width=\textwidth]{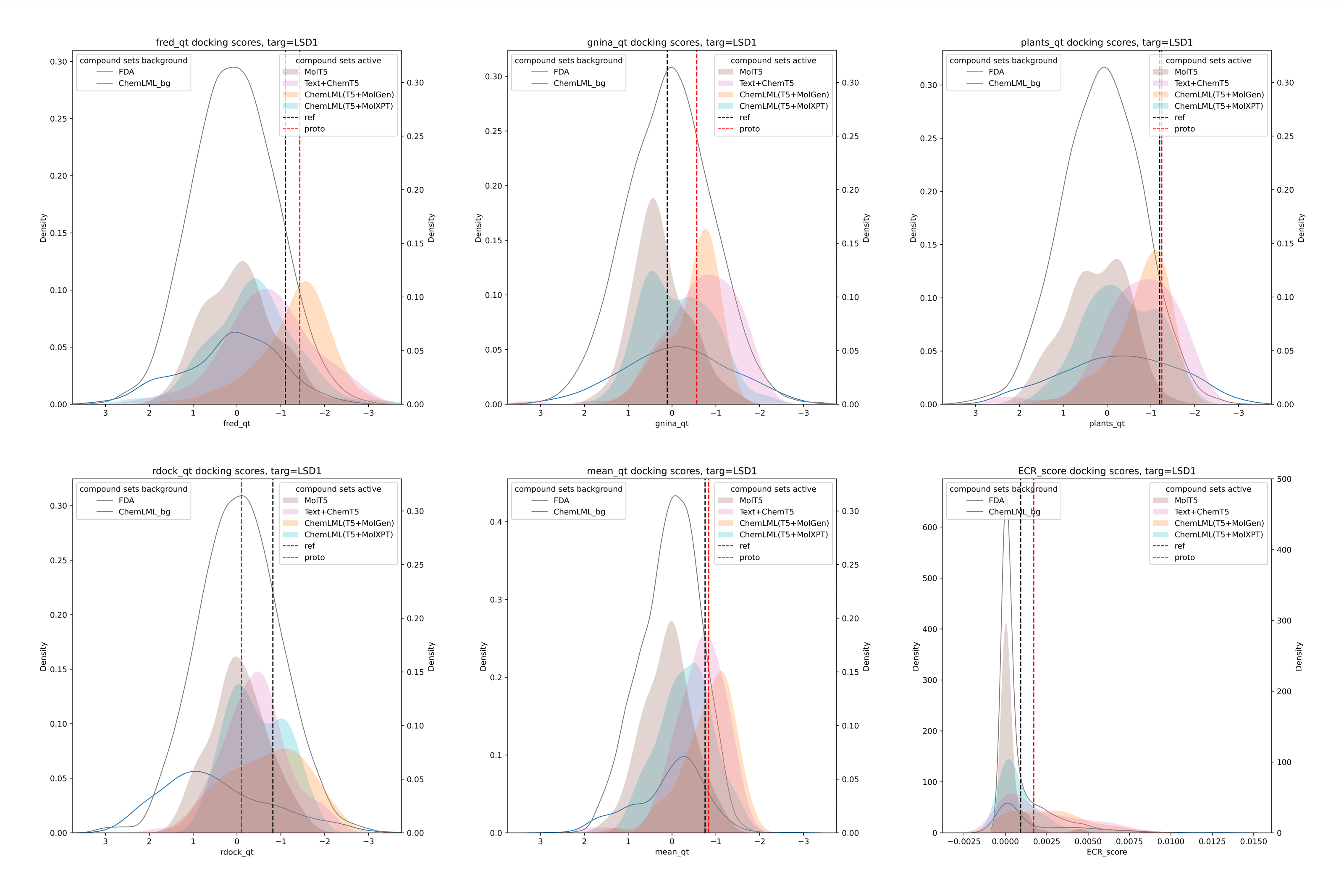}
    \caption{Individual docking program scores for LSD1.}
    \label{fig:individual_LSD1}
\end{figure}

\begin{figure}[htb]
    \centering
\includegraphics[width=\textwidth]{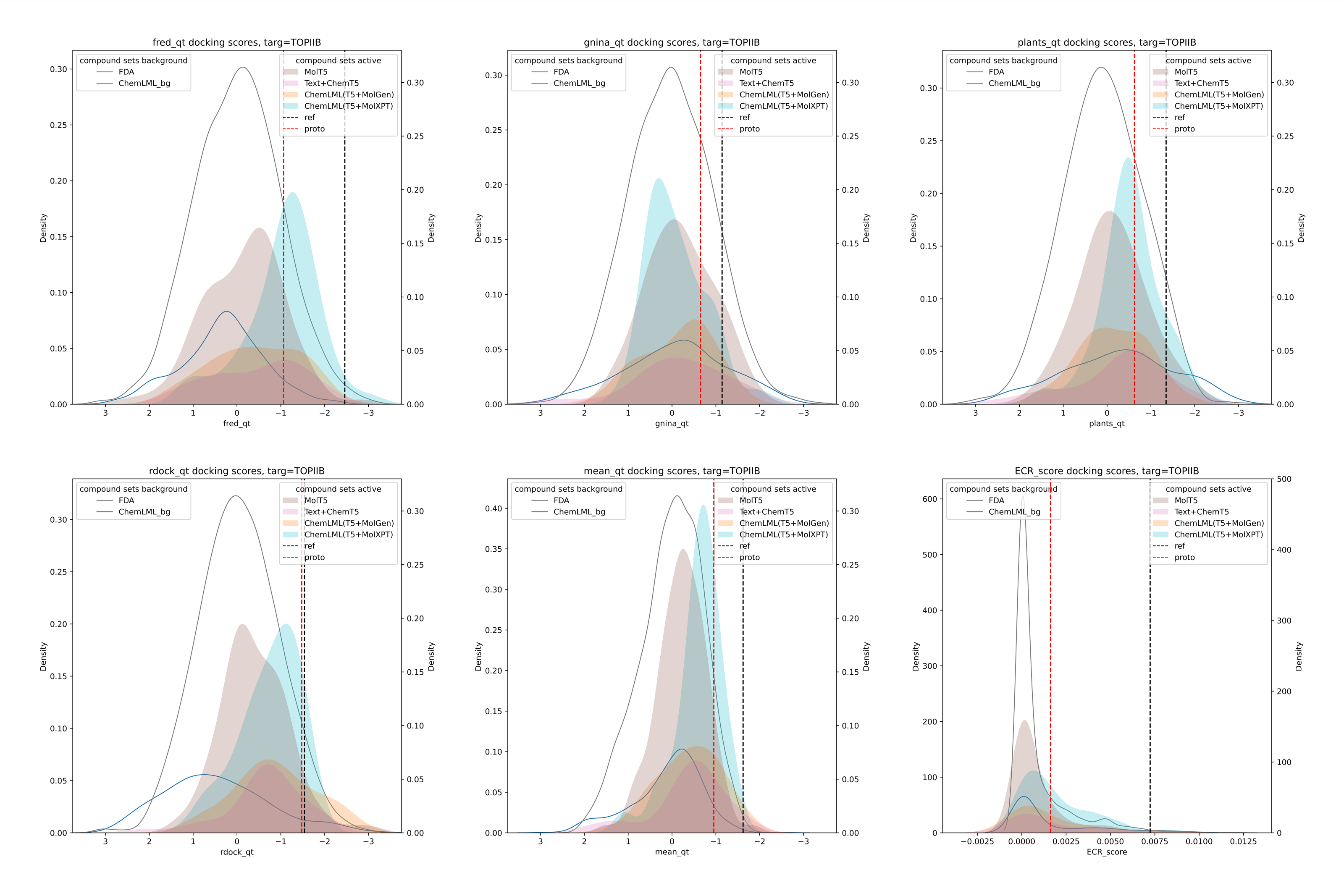}
    \caption{Individual docking program scores for TOPIIB.}
    \label{fig:individual_TOPIIB}
\end{figure}

\begin{figure}[htb]
    \centering
\includegraphics[width=\textwidth]{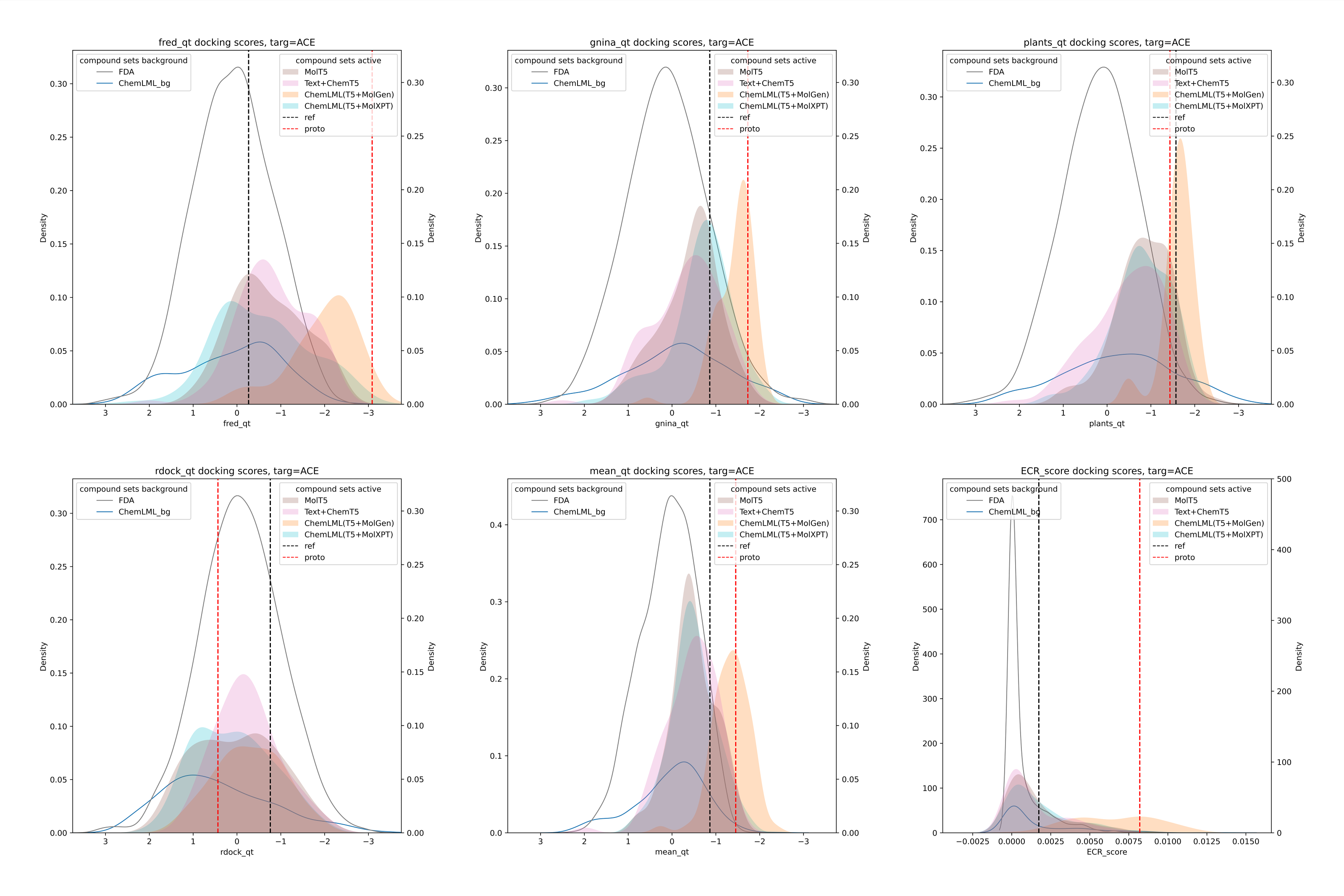}
    \caption{Individual docking program scores for ACE.}
    \label{fig:individual_ACE}
\end{figure}

\begin{figure}[htb]
    \centering
\includegraphics[width=\textwidth]{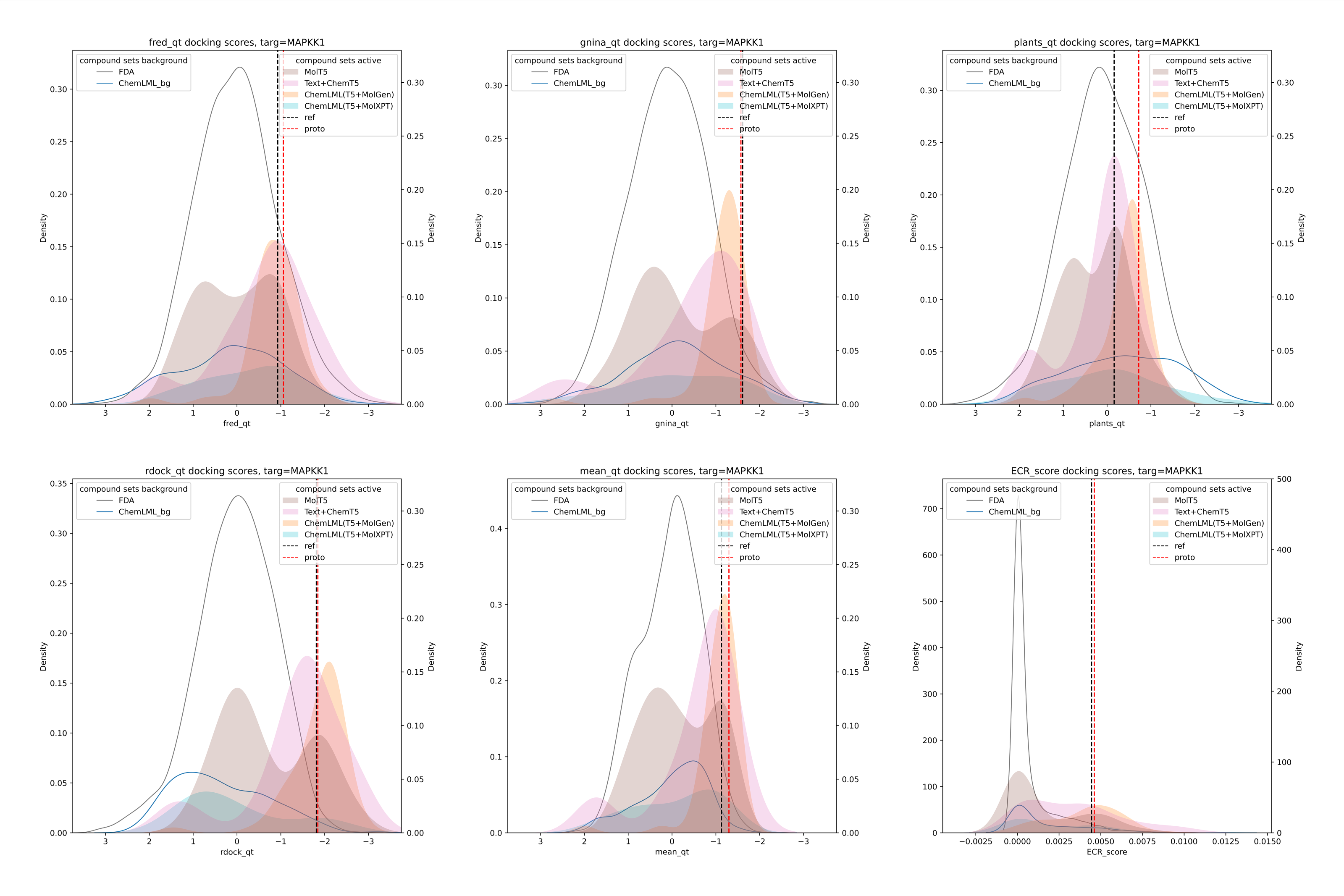}
    \caption{Individual docking program scores for MAPKK1.}
    \label{fig:individual_MAPKK1}
\end{figure}

To demonstrate how ChemLML could be used in practice, we applied it to generate chemical inhibitors of eight protein drug targets: acetylcholinesterase (AChE), inosine-5'-monophosphate dehydrogenase (IMPDH), heat shock protein HSP 90-alpha (HSP90AA1), SARS coronavirus main proteinase (M\textsuperscript{pro}), lysine-specific histone demethylase 1A (LSD1), DNA topoisomerase 2-beta
(TOPIIB), angiotensin-converting enzyme (ACE), and mitogen-activated protein kinase kinase 1 (MAPKK1). Details of the targets and text prompts are shown in Table~\ref{tab:docking}. % Using UniProt recommended names
We obtained the X-ray crystallographic structures of these proteins and used FRED \citep{mcgann2011FRED}, Gnina \citep{mcnutt2021gnina}, PLANTS \citep{korb2009PLANTS}, and rDock \citep{ruizcarmona2014rDock} to dock the ground truth inhibitor (co-crystallized or known high-affinity ligand) from PubChem, the ChemLML-generated candidate inhibitors, and candidate inhibitors generated by two baseline methods.
In addition, we docked control molecules generated by ChemLML and an FDA-approved screening library that are not expected to bind the targets.
We generated a single consensus docking score per compound (Figure~\ref{fig:docking_hist}).
Figures~\ref{fig:individual_AChE}-\ref{fig:individual_MAPKK1}
contain docking score distributions from the four individual docking programs.
They demonstrate the benefits of the consensus docking approach.
For example, on IMPDH, PLANTS and Gnina fail, but FRED, rDock, and the resulting consensus score succeed (Figure~\ref{fig:individual_IMPDH}).

In four of eight targets, the ChemLML(T5+MolGen) compounds dock with median scores exceeding those for their respective ground-truth ligand molecules. For IMPDH, the ground truth molecule, ribavirin monophosphate, and ChemLML(T5+MolGen) model score markedly better (0.00488 and median of 0.00594) than the control FDA and ChemLML background sets (medians of 0.00006 and 0.00049). ChemLML(T5+MolXPT), however, fails to improve over the backgrounds (0.00040).
The difference in median scores between the ChemLML variants could be related to the difference in valid 3D conformations produced by each of these ChemLML variants.
Median docking scores for ChemLML(T5+MolGen)-generated compounds for targets LSD1 (0.00249), ACE1 (0.00688), and MAPKK1 (0.00452) surpass those of their respective the ground truth compounds (0.00092, 0.00174, and 0.00445). In each case, the ground truth molecule and ChemLML(T5+MolGen) results show substantially better scoring than the control FDA and ChemLML background sets.

In two of the eight targets, AChE and TOPIIB, the ChemLML(T5+MolGen)-generated compounds score better than the control FDA and random ChemLML-generated sets but contain only a few compounds that exceed the ground truth molecules' scores. For AChE, the ChemLML(T5+MolGen)-generated molecules have docking scores (median 0.00237) well above the background distributions for FDA and generated sets (medians of 0.00013 and 0.00064) but below the ground truth molecule (0.00873). The other generated compounds from ChemLML(T5+MolXPT), MolT5, and Text+ChemT5 do not score better than the background sets. 
The first generated molecule from ChemLML(T5+MolGen) has a docking score (0.00482) above the background distributions, yet departs structurally from the ground truth molecule (Table \ref{tab:docking}), which is valuable in drug discovery. We also inspected whether the generated molecules satisfied the chemical structural properties in the text descriptions and their chemical structural diversity (Supplementary Results).
In the case of TOPIIB, the median docking scores for both ChemLML(T5+MolGen) (0.00052) and ChemLML(T5+MolXPT) (0.00124) both exceed those for the FDA (0.00007) and ChemLML-random (0.00011) background sets. However, the ground truth molecule (amsacrine--a DNA intercalator) docks more favorably (0.00723). 

ChemLML-generated compounds fair poorly against two of the eight targets, HSP90AA1 and M\textsuperscript{pro}. Docked ground truth and most generated molecules score poorly for HSP90AA1, likely due to poor handling of macrocycles (Table \ref{tab:docking}) by docking programs.
Here, control compounds from the FDA and ChemLML background sets have median scores (0.00009 and 0.00043, respectively) similar to the ground truth molecule (0.00006) and the ChemLML(T5+MolGen) median (0.00004). The ChemLML(T5+MolXPT) median scores slightly higher (0.00115) but not substantially better than the random ChemLML background. Considering the poor docking score of the ground truth molecule, we consider HSP90AA1 a fraught target for comparative performance evaluation of the molecule generators.
Also, we cannot draw conclusions about the quality of the 3D conformers built from the SMILES of the generated molecules.
In the case of M\textsuperscript{pro}, we see favorable scoring (0.00077) for the ground truth molecule (Savinin) but poor median scoring for both ChemLML models: T5+MolGen (0.00007) and T5+MolXPT (0.00006). 
The co-crystallized ligand for M\textsuperscript{pro} is a 5-mer peptide, which occupies a large substrate binding cavity, potentially conferring selectivity for larger (higher molecular weight) ligand compounds.  

During molecule generation, we also found a limitation of the models jointly trained on the mixture of natural language and chemicals. When we increase the temperature to produce diverse molecules, the models collapse and generate natural language text mixed with SMILES strings. This phenomenon worsens as molecule descriptions become less structured. We discuss this further in our permeability experiment below.

We selected one of the targets for which ChemLML(T5+MolGen) performed well, IMPDH, to assess how different components of the input text description contribute to the generation success.
We used ChemLML(T5+MolGen) to generate molecules using the full text description, the components of the description that pertain to the chemical structure, and the components that describe the molecular function, including the role as an IMPDH inhibitor.
With the molecular function prompt alone, generation performance is poor (Figure~\ref{fig:prompt_plot}).
The chemical structure prompt alone produces candidate inhibitors that are better than the control FDA and random ChemLML-generated sets with a median docking score of 0.00396.
The full prompt provides the best performance.
This limited-scope experiment suggests that the chemical structure is more important than the functional descriptions in our docking case study, but the functional descriptions do contribute some additional information.

% Background https://www.fda.gov/files/drugs/published/In-Vitro-Metabolism--and-Transporter--Mediated-Drug-Drug-Interaction-Studies-Guidance-for-Industry.pdf (page 9)
\subsection{Permeability case study}
Assessing a chemical's interactions with membrane transporters is an important part of drug development because those interactions can affect drug absorption, disposition, and excretion \citep{giacomini_membrane_2010}.
\textit{In vitro} membrane permeability is one chemical property assessed for this purpose, which can be measured with a MDR1-MDCK efflux ratio (ER) assay.
The assay determines whether a chemical is a substrate of MDR1, also known as P-glycoprotein, which controls drug export and is expressed in the brain endothelia, intestine, kidney, and liver \citep{giacomini_membrane_2010}.
A chemical is typically considered a substrate if the ER is $\geq 2$ \citep{giacomini_membrane_2010}.
We assess whether ChemLML and existing text-based molecule generation models can generate membrane permeable compounds.
Because the prompt (Methods) does not contain information about the MDR1 protein or the MDR1-MDCK ER assay, this case study serves as a more general test of the models' ability to generate compounds with complex properties relevant for drug development.

\begin{figure}
    \centering
    \includegraphics[width=0.9\linewidth]{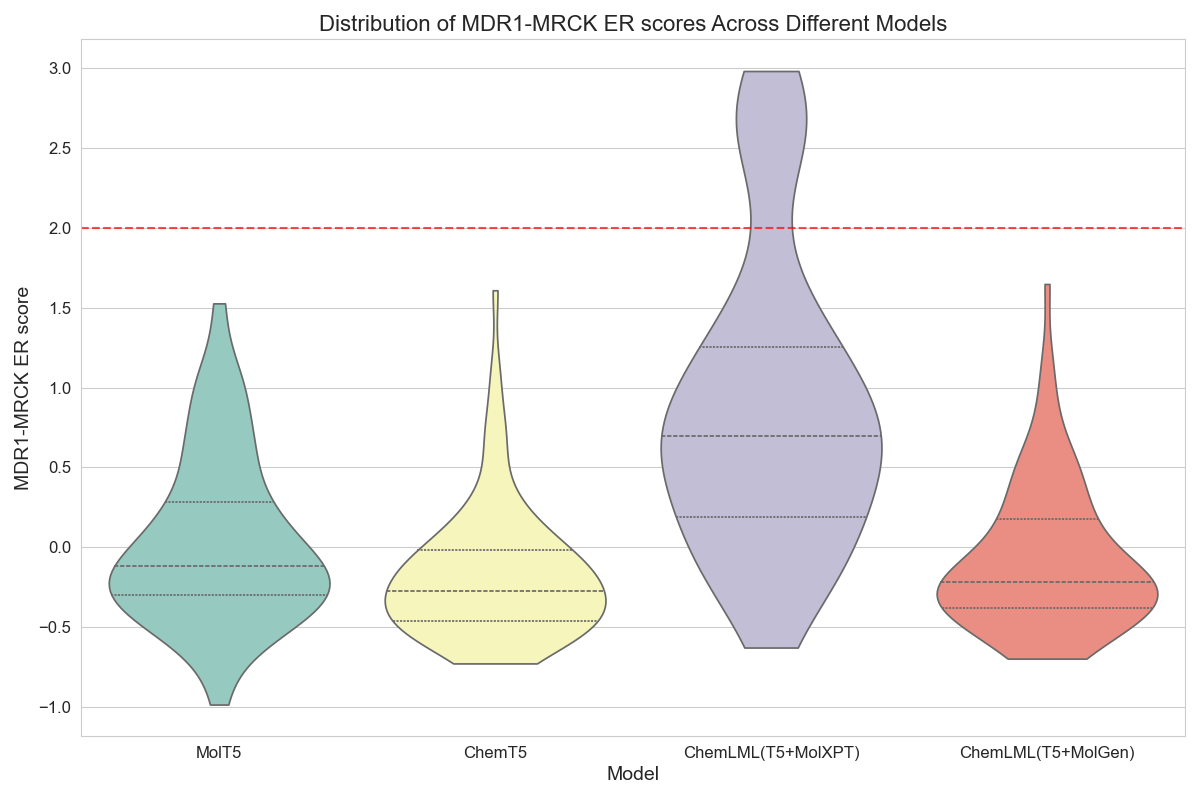}
    \caption{Distribution of MDR1-MRCK ER scores of different models. MDR1-MRCK ER score < 2 indicates predicted permeability.}
    \label{fig:MDR1-MRCK}
\end{figure}

We compare ChemLML(T5+MolXPT) and ChemLML(T5+MolGen) with the baselines MolT5 and Text+ChemT5.
All models run until they produce 100 unique molecules that pass multiple validity filters, and we assess how many duplicate molecules are generated and the percentage of unique molecules that pass these filters (Table~\ref{tab:permeability_count}). Then, we use an existing MDR1-MDCK ER model \citep{fang_prospective_2023} to predict the membrane permeability scores for these 100 generated molecules per model (Figure~\ref{fig:MDR1-MRCK}).
Because ER $\geq 2$ typically indicates the molecule is a MDR1 substrate, the generated permeable molecules should have a predicted MDR1-MRCK ER score $ < 2$.

All models generally succeed in generating membrane permeability molecules, and the distributions of predicted MDR1-MDCK ER scores are similar for three of the models.
ChemLML(T5+MolXPT) lags behind the other three models with a notable upward shift in the score distribution and some generated molecules that are predicted to be MDR1 substrates.
A possible explanation is that MolXPT itself is fine-tuned on ChEBI-20. We only use its molecule generation capability, thus it may not generalize well on this out-of-distribution description that occurs infrequently in the ChEBI-20 text.
In addition, we observe that ChemLML(T5+MolXPT) generates the highest number of duplicate molecules (Table~\ref{tab:permeability_count}), suggesting that it tends to converge on specific structures rather than generating diverse molecules.
Overall, ChemLML(T5+MolGen) demonstrates the highest success rate in molecule generation with the fewest filtered molecules and effectively produces molecules that are predicted to be permeable, as specified by the prompt.
By leveraging a single-modality molecular language decoder (MolGen), this ChemLML model avoids generating natural language tokens (like T5) while still effectively following natural language instructions.

\begin{table}[]
    \centering
    \begin{tabular}{c|c|c|c|c|c|c|c}
    \toprule
         Methods& Sample & Duplicate & Invalid & NL & Salts & SE & Success rate (\%) \\
         \hline
         MolT5& 260 & 31 & 113 & 8 & 4 & 4 &43.7\\
         Text+ChemT5& 984 & 28 & 318 & 395 & 101 & 42 & 10.6   \\
         ChemLML(T5+MolXPT)& 624& 519 & 3 & 1 & 1 & 0 & 95.2\\
         ChemLML(T5+MolGen)& 105& 3 &1 & 0 & 0 & 1 & 98.0 \\
    \bottomrule
    \end{tabular}
    \caption{Statistics of generated molecules for the permeability experiment. Each method generates molecules until it generates 100 successful molecules.
    \\ \textbf{Sample:} Number of total molecule generation sampling iterations
    \\ \textbf{Duplicate:} Number of duplicate molecules among the generated samples.\\$Sample - Duplicate = Unique$
    \\\textbf{Success rate (\%):} The percentage of molecules that successfully pass all four filters below relative to the number of unique generated molecules.  $100/Unique$
    \\These four cases are considered as an unsuccessful generation:
    \\\textbf{Invalid:} Number of generated molecules that could not be parsed by RDKit.
    \\\textbf{NL (natural language):} Number of molecules containing natural language text or tokens outside the molecular representation space.
    \\\textbf{Salts:} Number of molecules containing salt components.
    \\\textbf{SE (single element):} Number of molecules composed of only a single chemical element.
    \\$Unique$ equals the 100 successfully generated molecules plus the number of filtered molecules across these four categories.}
    
    \label{tab:permeability_count}
\end{table}

\section{Discussion}
This study proposes the ChemLML framework for combining pretrained natural language and molecular models in text-guided molecule design. By reusing large-scale pretrained models, we enhance the flexibility of text-based molecule generation and reduce the training effort by supporting multiple types of frozen text encoders.
In our case study generating candidate inhibitors for eight protein targets from text, we find that ChemLML T5 encoder+MolGen often generates molecules with docking scores that are better than ChemLML T5 encoder+MolXPT, two existing molecule generation algorithms, and two distributions of control molecules.
For four targets, the median ChemLML docking score is even better than the ground truth inhibitor from PubChem.

When finetuning LLMs, we find it is easy to finetune Galactica 125M. However, it becomes harder to finetune the Galactica 1.3B model.
Other finetuning methods such as Low-Rank Adaptation \citep{hu2022lora} may solve this problem. Also, we do not carry out experiments on Galactica 30B and 120B due to hardware and training technique limitations. Improving the LLM finetuning and finetuning larger LLMs are directions for future work.

A limitation of ChemLML is that it only focuses on molecule generation whereas related methods are multi-task. For instance, MolT5 can perform molecule captioning, and Text+ChemT5 can also conduct forward reaction and retrosynthesis prediction. Mol-Instruction \cite{fang2023mol} introduced multi-task dataset that supports a wide range of molecular learning tasks. Conceptually, the ChemLML framework is compatible with the multi-task setting and even more modalities including proteins \citep{liu2023text, liu_prott3_2024}. This presents another avenue for future work.

\begin{figure}[htb]
    \centering
\includegraphics[width=0.75\textwidth]{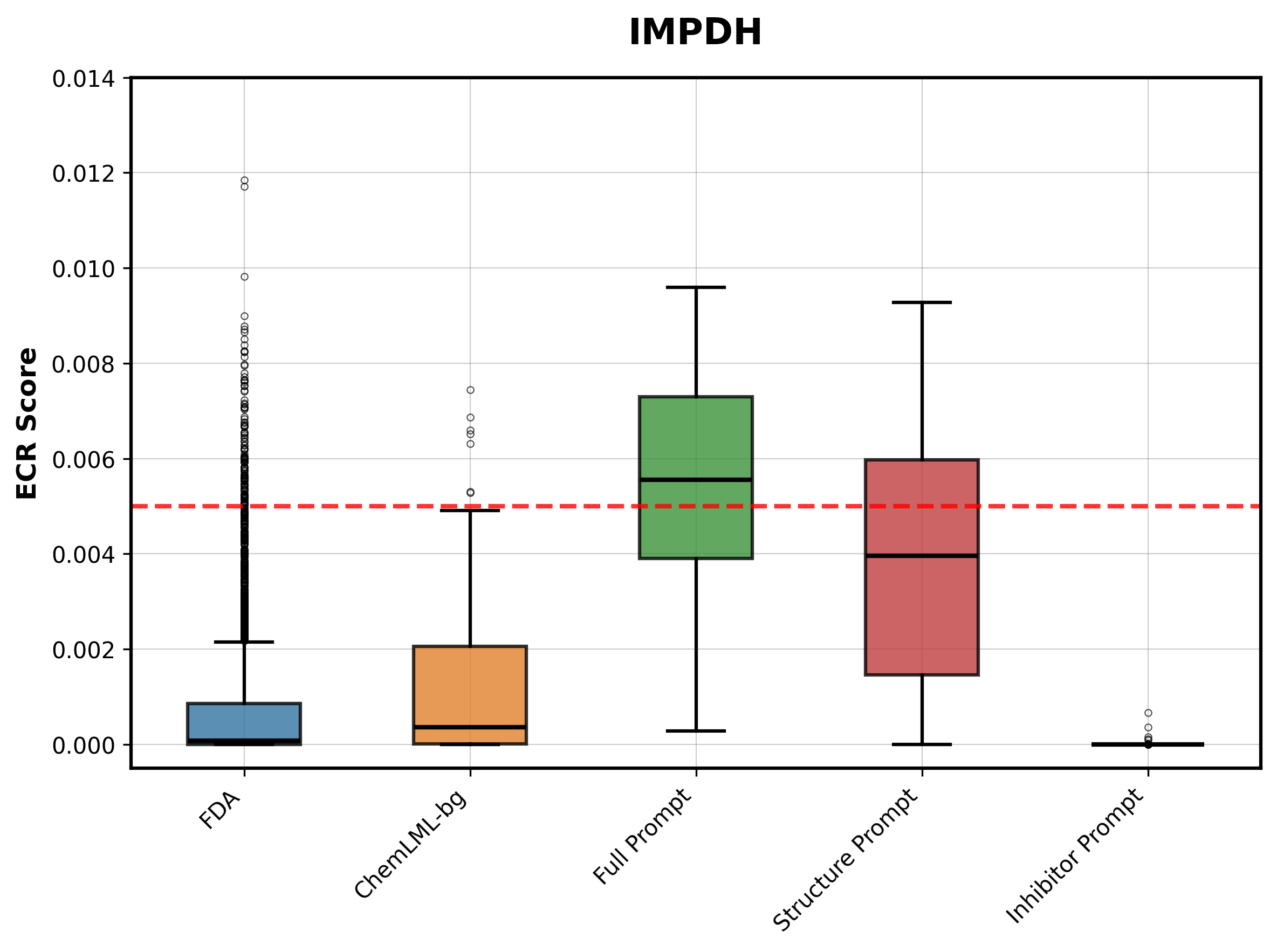}
    \caption{ChemLML(T5+MolGen) molecule generation with the IMPDH Full Prompt, Structure Prompt describing the chemical structure, or Inhibitor Prompt describing the molecular functions.}
    \label{fig:prompt_plot}
\end{figure}

Even though ChemLML performs well on the CheBI-20 and PubChem datasets, there is still a gap between these experiments and real world applications. The molecular descriptions in these datasets follow a relatively fixed format and can include both chemical structure and function information. We have not conducted user studies to assess how chemists' general expectations for text-based molecule generation deviate from the types of descriptions in CheBI-20 and PubChem.
In addition, a limitation of text-based molecule generators like ChemLML is that the text encoder can be sensitive to the prompt (Figure~\ref{fig:prompt_plot}).
The necessity to conduct evaluations on datasets like CheBI-20 and PubChem, where molecules have existing text descriptions, makes it challenging to evaluate these models' robustness and generalizability to out-of-distribution data, both new text prompts and new molecules.
Finally, our case study evaluations of the generated molecules rely on computational methods such as consensus docking and predicted MDR1-MDCK ER scores, which are themselves error-prone.
Certain types of ChemLML-generated molecules such as the candidate protein inhibitors could be assessed by more accurate computational methods such as absolute binding free-energy calculations 
\cite{ries_automated_2024, wu_optimizing_2025}.
However, there are no suitable predictive models for some chemical properties, and even the best computational models are still not a substitute for experimental validation.

Even after our PubChem filtering, data quality issues in this dataset likely remain (Supplementary Methods).
PubChem-unfiltered offers a large potential training dataset, and similar datasets have been used in prior work.
We intentionally decided to not train models on the generic text descriptions in PubChem and caution others to think carefully about whether PubChem-unfiltered or PubChem-filtered is more appropriate for their training and evaluation goals.

Many molecule generation models, including ChemLML, still face limitations in generating fully plausible, semantically-valid molecules even when then are syntactically-valid.
We observed that ChemLML-generated molecules in the SMILES or SELFIES format may fail to be transformed to 3D conformers for docking.
In general, inspection of other molecule generation methods has uncovered unstable ring systems, reactive or toxic moieties, or geometric errors \citep{walters2024generative}.
ChemLML may be susceptible to these issues as well.
Alternative molecular representations beyond SMILES and SELFIES \citep{ozcelik2025hitchhikers} could be relevant for addressing these issues in the future as well as additional careful assessments of the impact of the choice of molecular representation \citep{skinnider2024invalid}. 

Finally, generating structurally similar molecules does not necessarily ensure that they will share the same properties as the target molecule. Many molecule descriptions pertain to the chemical structure, i.e., bond topology or scaffold geometry, of the molecules.
In these cases, the assumption made by the chemical structure-based evaluations, like fingerprint similarity or docking, is most appropriate. For other physicochemical properties, like water solubility, different chemical structures could yield similar properties.
In these cases, the selection of improved evaluated metrics remains another area for future exploration. 

\section{Conclusion}
ChemLML introduces the strategy of using an adapter between text LLMs and molecule decoders for text-guided molecule design. Our approach is designed to capitalize on the rampant advances in both natural language modeling and unconditional molecule generation. The adapter is lightweight and compatible with different kinds of pretrained LLM encoders and molecule decoders.
Looking forward, the biological and chemical communities have shown strong interest in using natural language to condition the generation of small molecules and even proteins with desired structural and functional properties \cite{pei2024leveraging}.
ChemLML provides a path for combining the most advanced natural language and domain-specific models as they continue to undergo rapid development for this type of text-conditioned generation.

\section{Acknowledgments}
%\begin{ack}
This research was supported by National Institutes of Health awards R01GM135631 and P30CA014520. The research was performed using the computing
resources and assistance of the University of Wisconsin-Madison Center for High Throughput Computing \citep{https://doi.org/10.21231/gnt1-hw21}. We thank Shengchao Liu for addressing problems with dataset collection, Yin Fang for discussion of Mol-Instruction, and Chaowei Xiao for insightful discussion about the experimental setting.

\section{Author contributions}
% Loosely based on https://credit.niso.org/ but not using all of them
Y.D. and A.G. developed the ChemLML concept.
Y.D. implemented the ChemLML software and generated molecules.
S.S.E. performed the docking studies.
Y.D., S.S.E., and A.G. analyzed data and wrote the manuscript.

\section{Conflicts of interest}
The authors declare no competing financial interest.

\bibliography{references}

\clearpage
\beginsupplement
\noindent{\large \textbf{Supporting Information: Chemical Language Model Linker: blending text and molecules with modular adapters}}

\noindent Yifan Deng,$^{\dagger,\ddagger}$ Spencer S. Ericksen,$^{\P}$ Anthony Gitter$^{*,\S,\dagger,\ddagger}$

\vspace{1cm}

% ACS affiliation symbols from https://tex.stackexchange.com/questions/212227/how-do-i-remove-the-symbols-near-the-authors-names-in-achemso
\noindent $\dagger$ Department of Computer Sciences, University of Wisconsin-Madison, Madison, WI 53706, United States

\noindent $\ddagger$ Morgridge Institute for Research, Madison, WI 53715, United States

\noindent $\P$ Drug Development Core, Small Molecule Screening Facility, University of Wisconsin Carbone Cancer Center, University of Wisconsin-Madison, Madison, WI 53705, United States

\noindent $\S$ Department of Biostatistics and Medical Informatics, University of Wisconsin-Madison, Madison, WI 53792, United States

\noindent $^*$ Email: gitter@biostat.wisc.edu

\vspace{1cm}

%\noindent Pages: 23

%\noindent Figures: 10

%\noindent Tables: 3

\clearpage

% \mciteSetTracking{}

\clearpage
\subsection{Supplementary Figures}
\begin{figure}[htb]
    \centering    \includegraphics[width=\textwidth,height=\textheight,keepaspectratio]{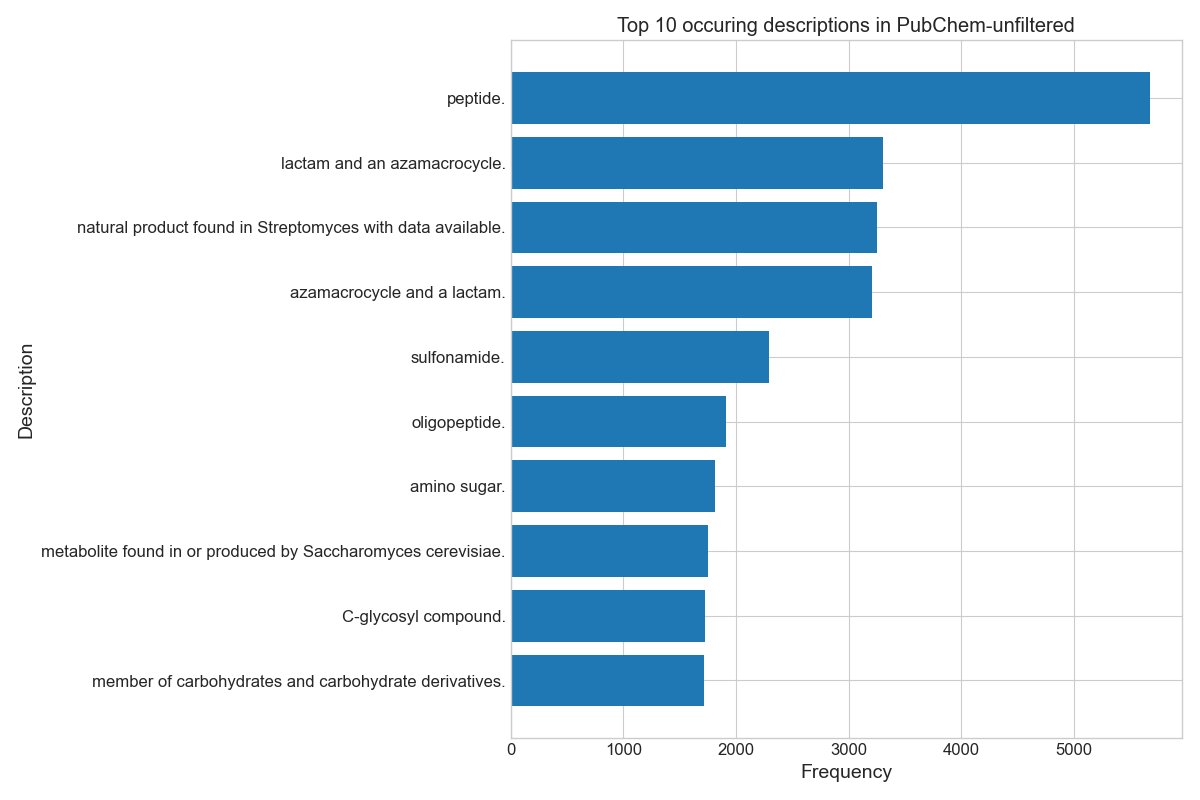}
    \caption{Top 10 most frequent descriptions from the PubChem-unfiltered dataset. The description prefix ``The molecule is a/an'' is omitted.}
    \label{fig:frequency}
\end{figure}

\clearpage
\subsection{Supplementary Tables}
\begin{table}[htb]
    \centering
\resizebox{\textwidth}{!}{%
\begin{tabular}{l|cccccccc}
\toprule

& Models & Trainable/Total Params& Exact $\uparrow$  &MACCS FTS $\uparrow$&RDK FTS $\uparrow$ & Morgan FTS $\uparrow$ & Validity $\uparrow$\\

\hline
\multirow{2}{*}{\parbox{2cm}{Baseline methods}}
&MolT5 & 248M/248M & 0.031 &0.475 &0.347 &0.278  &0.555 \\
& Text+ChemT5& 223M/223M  & 0.022  &  0.486 & 0.355 & 0.253 & 0.674 \\
\hline
\multirow{9}{*}{{ChemLML}}
&T5 encoder+MolXPT & 4.7M/464M & \textbf{0.031} & \textbf{0.474} & \textbf{0.347} & \textbf{0.224} &0.958  \\
&T5 encoder+MolGen 7B& 56.7M/6.64B & 0.028 & 0.446& 
0.302& 0.206& 0.992  \\
&SciBERT+MolGen & 4.7M/317M & 0.012 &  0.380 & 0.251 & 0.131 & 0.971\\
&Galactica 125M+MolGen & 4.7M/333M & 0.026 &0.414 &0.274 &0.177 & 0.990\\
&T5 encoder+MolGen & 4.7M/317M &0.012 & 0.408 & 0.260& 0.162& \textbf{0.995}\\
&Galactica 1.3B+MolGen & 7.4M/1.53B & 0.025 & 0.410 & 0.269 & 0.182 & 0.994\\ 
&Galactica 6.7B+MolGen & 11.5M/6.87B &0.024 & 0.423 & 0.270 & 0.189 & 0.992\\ 

\cline{2-8}
&T5 encoder finetune+MolXPT & 114M/464M & \textbf{0.034} &0.453&0.314& 
\textbf{0.229}& 0.988 \\
&SciBERT finetune+MolGen & 115M/317M & 0.027  &0.426 &0.289 &0.200 &0.990 \\
&T5 encoder finetune+MolGen & 114M/317M & 0.029 & 0.458&0.325&0.208&0.991\\
&Galactica 125M finetune+MolGen & 130M/333M &0.021 & 0.413 & 0.271 & 0.175 & 0.991 \\
&T5 encoder finetune+MolGen 7B & 166M/6.64B & 0.030  &\textbf{0.474} & \textbf{0.337}& 0.218& \textbf{0.995}\\
\hline
\end{tabular}
}
    \caption{Results of molecule generation on the PubChem-unfiltered test set.}
    \label{tab:PubChem_unfiltered_result}
\end{table}

\begin{table}[htb]
\footnotesize
\makebox[\textwidth][c]{
\begin{tabular}{l p{2cm} p{2cm} p{2cm} p{6cm} p{1.2cm} p{1.4cm}}
Similarity & Ground truth molecule (SMILES) & Ground truth molecule (PubChem CID) & Generated molecule (SMILES) & Description & Target protein (UniProt) & Target protein structure (PDB) \\
\hline
Low & {\scriptsize\seqsplit{C/C=C1\textbackslash{}\textbackslash{}[C@H{]}2C=C(C)C{[}C@{]}1(N)c1ccc(=O){[}nH{]}c1C2}} 
\adjustbox{right}{\includegraphics[width=0.25\textwidth]{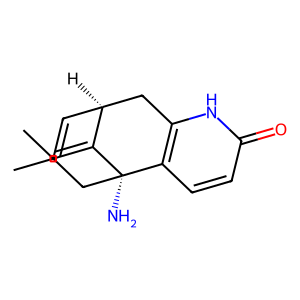}} & 854026 & {\scriptsize\seqsplit{CC1=CC{[}C@@H{]}(CCC{[}C@@H{]}(N)C=C2Cc3cccc(c3)N2){[}C@@H{]}1N}} \adjustbox{right}{\includegraphics[width=0.25\textwidth]{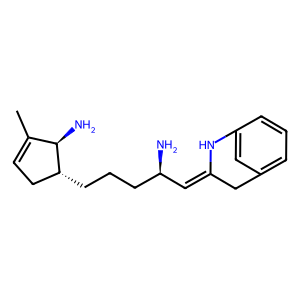}} & {\scriptsize The molecule is a sesquiterpene alkaloid isolated from a club moss Huperzia serrata that has been shown to exhibit neuroprotective activity. It is also an effective inhibitor of acetylcholinesterase and has attracted interest as a therapeutic candidate for Alzheimer's disease. It has a role as an EC 3.1.1.7 (acetylcholinesterase) inhibitor, a neuroprotective agent, a plant metabolite and a nootropic agent. It is a sesquiterpene alkaloid, a pyridone, a primary amino compound and an organic heterotricyclic compound. It is a conjugate base of a huperzine A(1+).} & P22303 & 4EY5 \citep{cheung2012acetylcholinesterase} \\
Medium & {\scriptsize\seqsplit{NC(=O)c1ncn({[}C@@H{]}2O{[}C@H{]}(COP(=O)(O)O){[}C@@H{]}(O){[}C@H{]}2O)n1}} \adjustbox{right}{\includegraphics[width=0.25\textwidth]{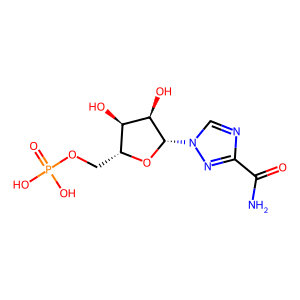}}& 100252 & {\scriptsize\seqsplit{NC(=O)c1cncc{[}n+{]}1N=C1N{[}C@H{]}(COP(=O)(O)O){[}C@@H{]}(O){[}C@H{]}1O}} \adjustbox{right}{\includegraphics[width=0.25\textwidth]{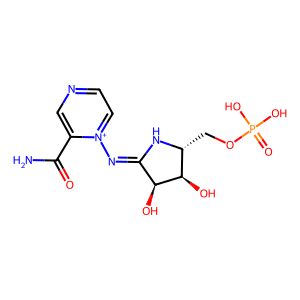}}& {\scriptsize The molecule is a 1-ribosyltriazole that is ribavirin in which the hydroxy group at the 5'-position is replaced by a phosphonooxy group. It is the active metabolite of the antiviral agent ribavirin. It has a role as a human blood serum metabolite, a drug metabolite, an antiviral agent and an EC 1.1.1.205 (IMP dehydrogenase) inhibitor. It is a 1-ribosyltriazole, a ribose monophosphate, an aromatic amide, a monocarboxylic acid amide and a primary carboxamide. It is functionally related to a ribavirin. It is a conjugate acid of a ribavirin 5'-monophosphate(2-).} & P50097 & 1ME7 \citep{prosise2002crystal} \\
High & {\scriptsize\seqsplit{CO{[}C@H{]}1/C=C\textbackslash{}\textbackslash{}C=C(/C)C(=O)NC2=CC(=O)C(NCCN(C)C)=C(C{[}C@@H{]}(C)C{[}C@H{]}(OC){[}C@H{]}(O){[}C@@H{]}(C)/C=C(\textbackslash{}\textbackslash{}C){[}C@@H{]}1OC(N)=O)C2=O}} \adjustbox{right}{\includegraphics[width=0.25\textwidth]{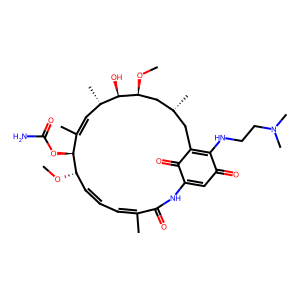}} & 5288674 & {\scriptsize\seqsplit{C=C{[}CH{]}C{[}C@@H{]}1CC(=O)C=C(C=O)C=CC(=O)NC(=O)/C(C)=C/C=C\textbackslash{}\textbackslash{}[C@H{]}(OC){[}C@@H{]}(OC(N)=O)/C(C)=C/{[}C@H{]}(C){[}C@@H{]}(O){[}C@@H{]}(OC)C1}} \adjustbox{right}{\includegraphics[width=0.25\textwidth]{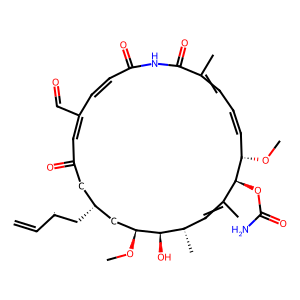}}& {\scriptsize The molecule is a 19-membered macrocyle that is geldanamycin in which the methoxy group attached to the benzoquinone moiety has been replaced by a 2-(N,N-dimethylamino)ethylamino group. It has a role as a Hsp90 inhibitor. It is a secondary amino compound, a tertiary amino compound, an ansamycin, a member of 1,4-benzoquinones and a carbamate ester. It is functionally related to a geldanamycin.} & P07900 & 1OSF \citep{jez2003crystal}
\end{tabular}
}
\caption{Text descriptions, ground truth molecules, generated molecules, and target proteins selected for the docking case study. The molecule columns also visualize the 2D structures.}
\label{tab:docking}
\end{table}
\clearpage

\begin{table}[htb]
\footnotesize
\makebox[\textwidth][c]{
\begin{tabular}{l p{2cm} p{2cm} p{2cm} p{6cm} p{1.2cm} p{1.4cm}}
Similarity & Ground truth molecule (SMILES) & Ground truth molecule (PubChem CID) & Generated molecule (SMILES) & Description & Target protein (UniProt) & Target protein structure (PDB) \\
\hline
High & {\scriptsize
\seqsplit{CCOC(=O){[}C@H{]}(CCc1ccccc1)N{[}C@@H{]}(C)C(=O)N1CCC{[}C@H{]}1C(=O)O.O=C(O)/C=C\textbackslash{}\textbackslash{}C(=O)O}
}  \adjustbox{right}{\includegraphics[width=0.25\textwidth]{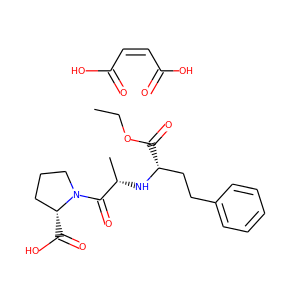}}
 & 5388961 & {\scriptsize
\seqsplit{C{[}C@H{]}(N{[}C@@H{]}(CCc1ccccc1)C(=O)O)C(=O)N1CCC{[}C@H{]}1C(=O)OOOC(Cc1ccccc1)C(=O)O}
} \adjustbox{right}{\includegraphics[width=0.25\textwidth]{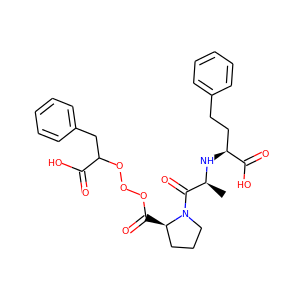}} & {\scriptsize The molecule is the maleic acid salt of enalapril. It contains one molecule of maleic acid for each molecule of enalapril. Following oral administration, the ethyl ester group of enalapril is hydrolysed to afford the corresponding carboxylic acid, enalaprilat, an angiotensin-converting enzyme (ACE) inhibitor. Enalapril is thus a prodrug for enalaprilat (which, unlike enalapril, is not absorbed by mouth), and its maleate is used in the treatment of hypertension and heart failure, for reduction of proteinuria and renal disease in patients with nephropathies, and for the prevention of stroke, myocardial infarction, and cardiac death in high-risk patients. It has a role as an EC 3.4.15.1 (peptidyl-dipeptidase A) inhibitor, an antihypertensive agent and a prodrug. It contains an enalapril.} & P12821  & 1UZE \citep{natesh_structural_2004} \\
High & {\scriptsize
\seqsplit{O=C1OC{[}C@H{]}(Cc2ccc3c(c2)OCO3)/C1=C\textbackslash{}\textbackslash{}c1ccc2c(c1)OCO2}
} \adjustbox{right}{\includegraphics[width=0.25\textwidth]{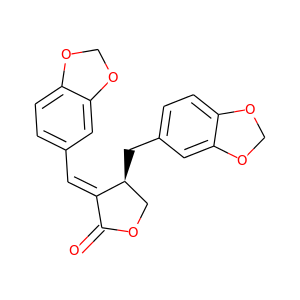}} & 5281867 & {\scriptsize
\seqsplit{O=C1OC{[}C@H{]}(Cc2ccc3c(c2)OCO3){[}C@H{]}1Cc1ccc2c(c1)OCO2}
} \adjustbox{right}{\includegraphics[width=0.25\textwidth]{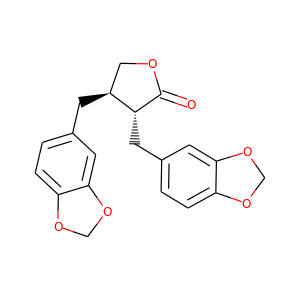}} & {\scriptsize The molecule is a lignan that is dihydrofuran-2(3H)-one (gamma-butyrolactone) substituted by a 1,3-benzodioxol-5-ylmethylidene group at position 3 and a 1,3-benzodioxol-5-ylmethyl group at position 4 (the 3E,4R-isomer). It exhibits antiviral activity against SARS-CoV-2. It has a role as a T-cell proliferation inhibitor, an anti-inflammatory agent, a plant metabolite, an EC 3.4.22.69 (SARS coronavirus main proteinase) inhibitor and an anticoronaviral agent. It is a member of benzodioxoles, a lignan and a gamma-lactone.} & P0C6X7 & 1UK4 \citep{doi:10.1073/pnas.1835675100} \\
High & {\scriptsize
\seqsplit{CCO.N\#CC(=C(/N)Sc1ccccc1N)/C(C\#N)=C(\textbackslash{}\textbackslash{}N)Sc1ccccc1N}} \adjustbox{right}{\includegraphics[width=0.25\textwidth]{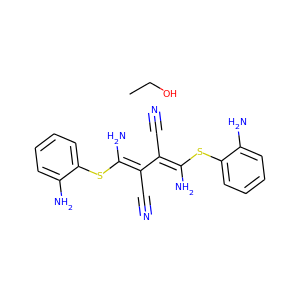}}  & 16220066 & {\scriptsize
\seqsplit{CCON=CC(=C(/N)Sc1ccccc1N)/C(C\#N)=C(\textbackslash{}\textbackslash{}N)Sc1ccccc1N}} \adjustbox{right}{\includegraphics[width=0.25\textwidth]{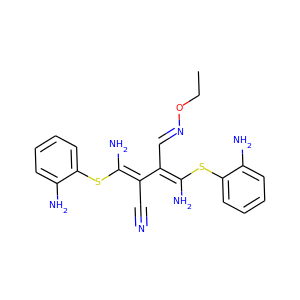}} & {\scriptsize The molecule is an addition compound obtained by combining equimolar amounts of (2Z,3Z)-bis{amino[(2-aminophenyl)sulfanyl]methylidene}butanedinitrile (U0126) and ethanol. An inhibitor of mitogen-activated protein kinase that also exhibits anti-cancer properties. It has a role as an EC 2.7.11.24 (mitogen-activated protein kinase) inhibitor, an apoptosis inducer, an antineoplastic agent, an antioxidant, an osteogenesis regulator and a vasoconstrictor agent. It contains an U0126.} & Q02750 & 3EQH \citep{doi:10.1021/bi801898e} \\
\end{tabular}
}
\label{tab:docking_1}
\end{table}

\clearpage

\begin{table}[htb]
\footnotesize
\makebox[\textwidth][c]{
\begin{tabular}{l p{2cm} p{2cm} p{2cm} p{6cm} p{1.2cm} p{1.4cm}}
Similarity & Ground truth molecule (SMILES) & Ground truth molecule (PubChem CID) & Generated molecule (SMILES) & Description & Target protein (UniProt) & Target protein structure (PDB) \\
\hline
Low & {\scriptsize
\seqsplit{CC(O)C(=O)O.COc1cc(NS(C)(=O)=O)ccc1Nc1c2ccccc2nc2ccccc12}} \adjustbox{right}{\includegraphics[width=0.25\textwidth]{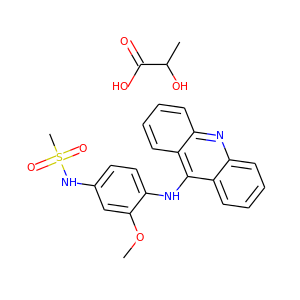}} & 88124 & {\scriptsize\seqsplit{NC=CC=CC1=CCc2cc(N)cc3c2c(=O)n1c1ccccc31}} \adjustbox{right}{\includegraphics[width=0.25\textwidth]{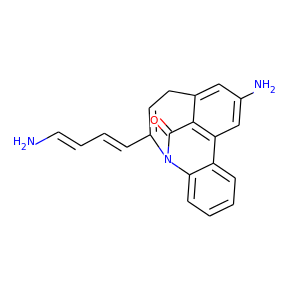}} & The molecule is the lactate form of amsacrine, an aminoacridine analog and topoisomerase II inhibitor, with antineoplastic activity. Although the exact relationship between DNA binding and its activity has yet to be fully elucidated, amsacrine intercalates DNA through its acridine moiety, and its nonintercalative headgroup impedes topoisomerase II activity, augmenting enzyme-mediated DNA cleavage and resulting in DNA double-strand breaks. This ultimately induces programmed cell death.& Q02880 & 4G0U \citep{10.1093/nar/gkt828} \\
Medium & {\scriptsize\seqsplit{O=C(O)c1ccc(CN2CCC(CN{[}C@@H{]}3C{[}C@H{]}3c3ccccc3)CC2)cc1}} \adjustbox{right}{\includegraphics[width=0.25\textwidth]{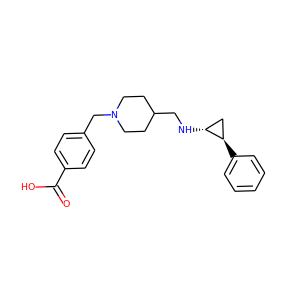}} & 66571643 & {\scriptsize\seqsplit{O=C(O)c1ccc(CNCCC2CCC2(NCCC2CC2)c2ccccc2)cc1}} \adjustbox{right}{\includegraphics[width=0.25\textwidth]{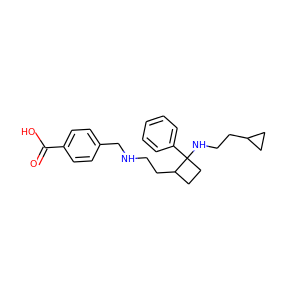}} & The molecule is a member of the class of piperidines that is piperidine substituted by (4-carboxyphenyl)methyl and {[(1R,2S)-2-phenylcyclopropyl]amino}methyl groups at positions 1 and 4, respectively. It is a potent and irreversible inhibitor of lysine specific demethylase 1 (LSD1, also known as KDM1A). It was under clinical investigation for the treatment of acute myeloid leukaemia and small cell lung carcinoma. It has a role as an EC 1.14.99.66 (lysine-specific histone demethylase 1A) inhibitor and an antineoplastic agent. It is a member of benzoic acids, a monocarboxylic acid, a member of piperidines, a member of cyclopropanes, a tertiary amino compound, a secondary amino compound and a member of benzenes. & O60341 & 6NQU \citep{10.3389/fimmu.2019.01351} \\
\end{tabular}
}
\label{tab:docking_2}
\end{table}

\clearpage

\begin{sidewaystable}[]
\begin{tabular}{cccccccccc}
 & \multicolumn{1}{l}{} & \multicolumn{8}{c}{Counts (N)} \\ \cline{3-10} 
Molecule set & Stage & HSP90AA1 & ACE1 & M\textsuperscript{pro} & MAPKK 1 & IMPDH & LSD1 & AChE & TOPIIB \\ \hline
 &  &  &  &  &  &  &  &  &  \\ \hline
\multirow{3}{*}{FDA} & docking & 2956 & 2956 & 2956 & 2956 & 2954 & 2956 & 2952 & 2956 \\
 & omega & 2977 &  &  &  &  &  &  &  \\
 & download & 3105 &  &  &  &  &  &  &  \\ \hline
\multirow{3}{*}{ChemLML background} & docking & 657 & 657 & 657 & 657 & 657 & 657 & 657 & 657 \\
 & omega & 663 &  &  &  &  &  &  &  \\
 & generation & 1000 &  &  &  &  &  &  &  \\ \hline
\multirow{3}{*}{ChemLML(T5+MolGen)} & docking & 16 & 66 & 43 & 54 & 42 & 75 & 52 & 44 \\
 & omega & 16 & 66 & 43 & 54 & 42 & 75 & 52 & 45 \\
 & generation & 100 & 100 & 100 & 100 & 100 & 100 & 100 & 100 \\ \hline
\multirow{3}{*}{ChemLML(T5+MolXPT)} & docking & 99 & 92 & 95 & 28 & 99 & 95 & 92 & 90 \\
 & omega & 99 & 92 & 95 & 28 & 99 & 95 & 92 & 90 \\
 & generation & 100 & 100 & 100 & 28 & 100 & 100 & 100 & 100 \\ \hline
\multirow{3}{*}{MolT5} & docking & 97 & 97 & 98 & 89 & 98 & 99 & 96 & 97 \\
 & omega & 97 & 97 & 98 & 89 & 99 & 99 & 96 & 97 \\
 & generation & 100 & 99 & 100 & 91 & 100 & 99 & 100 & 100 \\ \hline
\multirow{3}{*}{Text+ChemT5} & docking & 83 & 98 & 94 & 94 & 96 & 97 & 86 & 31 \\
 & omega & 83 & 98 & 94 & 94 & 99 & 98 & 86 & 31 \\
 & generation & 100 & 100 & 100 & 100 & 100 & 100 & 100 & 34 \\ \hline
\end{tabular}
\\
    \caption{Attrition of model-generated SMILES for each target from generation to docking output. The values indicate the number of SMILES successfully generated (RDKit-canonicalized); OpenEye (OE) canonicalized, protonated, and embedded into 3D coordinates by Omega2; and ultimately docked.}
    \label{tab:Generated_Cpd_Attrition}
\end{sidewaystable}

\clearpage

\subsection{Supplementary Methods}
\subsubsection{Computational resources}
For ChemLML models with less than 1B total parameters, we perform the training on one NVIDIA RTX 2080Ti, which has 11GB memory; for ChemLML models with greater than 1B total parameters, we perform the training on one NVIDIA L40, which has 48GB memory.

\subsubsection{PubChem dataset}
Generic PubChem descriptions may be reasonable for learning general molecule properties but violate our assumptions for evaluating text-based molecule generation, which leverages the molecular similarity principle and assumes the generated molecule should be structurally similar to the ground truth molecule.
The biggest problem with PubChem-unfiltered is that many molecule descriptions are general. For example, the most frequent description, occurring 5,753 times is ``The molecule is a peptide'', which is uninformative. The top 10 most frequent descriptions are shown in Figure \ref{fig:frequency}.  Also, there are descriptions like ``The molecule is a natural product found in'' when there can be hundreds of natural products produced by a single species.
For instance, the fungal genera \textit{Aspergillus} and \textit{Penicillium} are associated with 3,091 and 2,550 natural products, respectively, and the bacterial genus \textit{Streptomyces} is associated with 5,755 natural products \citep{van2022natural}. % https://doi.org/10.1093/nar/gkab941
% Counts from graphs at https://www.npatlas.org/discover/overview
% Example natural product entry https://www.npatlas.org/explore/compounds/NPA018564
% Could also use COCONUT or LOTUS data but cannot find sources online, would have to download and count locally

In addition, there are counterintuitive descriptions. One PubChem-unfiltered example has the description ``The molecule is a mineral'' and the SMILES of oxygen. It turns out that ``minerals'' is indeed in the PubChem page of oxygen (CID 977). This is because ``orange'' $^{18}O_2$ was crystallized under special experimental conditions \citep{cromer1983oxygen}.
Thus, oxygen is recorded in the American Mineralogist Crystal Structure Database, which is cross-referenced from PubChem. In order to obtain a more meaningful dataset, we filter out descriptions less than 30 words and containing the description ``natural product''.
This eliminates many, but not all, of the issues described above.

Even the PubChem-filtered dataset contains other potential problems.
A form of data leakage arguably occurs when a synonym of the molecule is in the text description.
Our dataset review noted many drug names such as loperamide, tretinoin, lanreotide, ambrisentan, nesbuvir, danusertib, and others in the descriptions.
Other descriptions state that the molecule is an enantiomer of another.
%, chemical branch names and synonyms. 
Yet other descriptions retain leftover IUPAC names, which can be used to generate molecules \citep{rothchild2021c5t5}.
In all of these cases, the text descriptions gives more information than expected about the chemical structure, which may inflate similarity-based evaluation metrics.

We constructed the PubChem-filtered dataset independently from the ChEBI-20 dataset.
However, even though we transformed all the molecules into canonical form and made sure there were no overlapping canonical SMILES between the PubChem-filtered and the ChEBI-20 datasets, there were still molecules that have 100\% similarity between PubChem-filtered and ChEBI-20. We did not exhaustively remove these molecules in the datasets.
Thus, we cannot guarantee that PubChem-filtered is entirely disjoint from the ChEBI-20 training and validation splits.
% PubChem has ChEBI as a source we believe

In the evaluations, we directly test MolT5 and Text+ChemT5's performance on the test set without further finetuning. For fair comparison, we also train ChemLML models on ChEBI-20 and test on the PubChem test set without finetuning. To make sure the PubChem test set corpus aligns with the ChEBI-20 training corpus, we manually replace each molecule's IUPAC name with ``This molecule'' and delete ``with data available'' at the end of the sentence.

Note that the typo ``macrocyle'' in the high similarity description in Table~\ref{tab:docking} appears in the original PubChem data and was not introduced by our processing.

\subsubsection{Evaluation metrics}
Fréchet ChemNet Distance \citep{preuer2018frechet} is a common evaluation metric for generative molecular models. It uses the embedding of the molecules in ChemNet \citep{mayr2018ChemNet}, a long short-term memory-based model, to detect whether generated molecules are diverse and have similar properties as real molecules. We do not use this metric because our preliminary results and an independent evaluation \citep{holzgruber2024GuacaMolEval} found it is highly sensitive to the sample size and molecule padding length.

Previous work in molecule generation also used metrics such as diversity and novelty for evaluating molecule generation. However, these metrics primarily focus on sampling the chemical space to generate diverse molecules, which are not well-suited for description-guided molecule design.

Levenshtein distance and BiLingual Evaluation Understudy (BLEU) scores have also been used previously to compare string representations of molecules. However, these two scores that are common in NLP are not as suitable for molecules. Imagine the case where the generated molecule exactly matches the ground truth molecule at every character except the last right parenthesis, which makes it fail to close the aromatic ring. It will a yield high BLEU score and low Levenshtein distance, despite the generated molecule being chemically invalid.

\subsubsection{Text+ChemT5 tasks}
The Text+ChemT5 \citep{christofidellis2023unifying} tasks are:
\begin{itemize}
    \item mol2mol: This task contains forward reaction and retrosynthesis subtasks. In the forward reaction task, given reagents and/or enzymes, the model needs to generate the main product of the chemical reaction. For retrosynthesis, given the product of a chemical reaction, the model needs to find the reagents and/or enzymes.
    \item mol2text: Given the molecule represented as SMILES, the model needs to generate a textual description of the molecule.
    \item text2mol: Given the textual description of a molecule, the model needs to generate the SMILES representation the molecule.
    \item text2text: Given the natural language description of a chemical reaction, the model needs to generates a step-wise execution protocol to carry out the reaction.
\end{itemize} 

\subsubsection{Docking case study}
To select targets from each similarity bin (low, medium, and high), we manually reviewed the instances in each bin.
We selected the first five descriptions that pertained to a single, specific protein target as opposed to multiple targets or inhibition of a biological process.
Then, we examined the ground truth molecules in PubChem.
We preferred experimental ligand-bound structures cross-referenced from PubChem with unambiguous binding sites for search space specification in docking.
In some cases, there was not a suitable ligand-bound structure directly cross-referenced from PubChem, but we were able to find a suitable structure in the RCSB Protein Data Bank \citep{burley_rcsb_2023}.
We used DrugBank \citep{knox_drugbank_2024} to confirm protein targets of the ground truth molecules as needed.
Table~\ref{tab:docking} provides more information about the ground truth molecules and target proteins.

For each target protein, we docked the ground truth molecule, generated molecules, and control molecules from two types of background distributions.
The first background distribution originally contained 3,082 small molecules (3,105 total substances) from the Selleck L1300 FDA-approved Library (downloaded 2024-03-25).  
The second background distribution originally contained 1,000 generated molecules from the ChemLML(T5+MolGen) model sampling text descriptions from PubChem-filtered.  
Due to molecule pre-processing, docking-based scoring was ultimately achieved on 2,956 and 657 compounds from these background sets, respectively (Table~\ref{tab:Generated_Cpd_Attrition}).
Additional components within generated SMILES, like complexed waters or counterions on salts, were stripped from parent molecular species prior to RDKit SMILES canonicalization.
Using OpenEye applications (Cadence Molecular Sciences, Santa Fe, NM), small molecule sets were processed from SMILES inputs into 3D conformers using Omega2-v5.0.0.3 and assigned partial charges (MMFF) using MolCharge from QUACPAC v2.2.3.3 \citep{hawkins2010omega}.
The most common error in 3D conformer generation involved missing force field parameters. 

Protein target structures for docking were downloaded from RCSB.org\citep{burley_rcsb_2023}: HSP90AA1 (PDB: 1OSF), IMPDH (1ME7), AChE (4EY5), ACE (1UZE), M\textsuperscript{pro} (1UK4), MAPKK1 (3EQH), LSD1 (6TE1), and TOPIIB (4G0U)  in PDB format and processed using the DockPrep utility in ChimeraX v1.7.1 \citep{meng2010chimerax}.
Compounds were docked with four different docking programs: FRED 4.3.0.3 (Cadence Molecular Sciences), Gnina v1.1 (\url{https://github.com/gnina/gnina}), PLANTS v1.2 (\url{https://github.com/discoverdata/parallel-PLANTS}), and rDock v24.03.192 (\url{https://github.com/CBDD/rDock}).

Docking site locations on each target were specified based on the position of the co-crystallized ground truth molecule in each target's protein crystal structure.
Docking scores for each molecule, for each of the four programs, were integrated for a consensus score based on the method of exponential consensus ranking as described in \citet{palaciorodriguez2019exponential}.

As a small control study, we assessed the effect of varying the text prompt for IMPDH inhibitor generation.
The full prompt is the entire description in Table~\ref{tab:docking}.
The structure prompt contains the parts of the description that pertain to chemical structure: ``The molecule is a 1-ribosyltriazole that is ribavirin in which the hydroxy group at the 5'-position is replaced by a phosphonooxy group. It is a 1-ribosyltriazole, a ribose monophosphate, an aromatic amide, a monocarboxylic acid amide and a primary carboxamide. It is a conjugate acid of a ribavirin 5'-monophosphate(2-)''.
The inhibitor prompt contains the parts of the description that describe the functional role as an IMPDH inhibitor: ``The molecule has a role as a human blood serum metabolite, a drug metabolite, an antiviral agent and an EC 1.1.1.205 (IMP dehydrogenase) inhibitor''.

\clearpage
\subsection{Supplementary Results}
\subsubsection{Broader impacts}
Like all molecule generation models, ChemLML has both positive and negative potential broader impacts due to its potential to suggest both beneficial and harmful novel chemicals \citep{urbina2022dual}.
These impacts are especially pertinent for text-based generative models such as ChemLML because they are designed to produce chemicals with desired properties from natural language without requiring experimental training data related to those properties for supervised training.
As demonstrated in our case studies, one goal with ChemLML is to use it for beneficial purposes through applications in drug discovery and development.
However, we also see the potential to generate harmful molecules based on our partial manual review of the ChEBMI-20 dataset used to train ChemLML models.
We encountered text descriptions related to opioids and carcinogens.
Ultimately, molecules generated by ChemLML still have to be manually reviewed and synthesized by a human chemist.
Therefore, we believe it presents less relative risk than existing harmful chemicals or fully-automated systems like ChemCrow \citep{bran2024ChemCrow} that plan and execute chemical synthesis.

\subsubsection{Generated molecule validity}
Not all ChemLML-generated molecules were chemically valid (Table~\ref{tab:Generated_Cpd_Attrition}).
Unlike other evaluations of generative molecule models that use RDKit’s molecular structure parser \cite{polykovskiy2020molecular}, we required that a molecule can be processed with RDKit and Omega2 as well as docked in order to be valid.
We examined the output from these steps to assess why generated molecules were invalid.
Most molecules were rejected during Omega2 processing rather than RDKit parsing or docking, often as a result of missing forcefield parameters.
The greatest attrition occurred due to the challenge of building macrocycles from generated SMILES for the HSP90AA1 target.
Despite running Omega2 in ``macrocycle'' mode, only 16 of 100 syntactically valid SMILES (mostly macrocycles) generated by ChemLML(T5+MolGen) for this target produced dockable structures.

\subsubsection{Docking case study generated molecules}
The descriptions of the IMPDH and HSP90AA1 inhibitor examples include the chemicals' functional roles as protein inhibitors as well as specific chemical structural attributes.
We examined the first generated molecules generated with temperature 1 and random seed 42, shown in Table~\ref{tab:docking}, to assess whether these structural attributes were present.
The structure of the ground truth IMPDH inhibitor is in part described as ``ribavirin in which the hydroxy group at the 5'-position is replaced by a phosphonooxy group''.
The corresponding generated molecule does contain the phosphonooxy (phosphate) group, but the 1,4-anhydroerythritol substructure in the ground truth molecule has an O replaced with NH in the generated molecule. % https://pubchem.ncbi.nlm.nih.gov/compound/641773 and https://pubchem.ncbi.nlm.nih.gov/compound/5460607
In the HSP90AA1 inhibitor example, ChemLML generates a macrocycle, but it is a 22-memberered macrocycle instead of a 19-membered macrocycle.
The generated molecule lacks the 1,4-benzoquinone substructure from the description but does contain the carbamate ester substructure.
Even in the AChE inhibitor example, where specific details on structure are absent in the description text, key unsaturated bicyclo(3.3.1)nonyl substructures are present in both the ground truth and generated compounds.
In these limited examples, ChemLML retains some structural properties in the generated molecules but omits others.
Furthermore, when we cluster the molecules that ChemLML generates for each target, we observe that they reflect diverse chemical structures. 

\end{document}